\documentclass[acmsmall]{acmart}
\usepackage{colortbl, xcolor}
\usepackage{booktabs,caption,fixltx2e}
\usepackage[flushleft]{threeparttable}

\usepackage{cellspace}
\usepackage{makecell} 
\newcommand\thickhline{\Xhline{0.08em}}

\usepackage{array}
\newcolumntype{P}[1]{>{\centering\arraybackslash}p{#1}}
\newcolumntype{?}{!{\vrule width 0.7pt}}
\usepackage{multirow}
\usepackage{diagbox}
\usepackage{amsmath}
\usepackage{graphicx}
\AtBeginDocument{%
  \providecommand\BibTeX{{%
    \normalfont B\kern-0.5em{\scshape i\kern-0.25em b}\kern-0.8em\TeX}}}

\begin{document}

\title{Toward Personalized Affect-Aware Socially Assistive Robot Tutors in Long-Term Interventions for Children with Autism}

% Long-Term, Multi-Session Modeling Affect of Student's with Autism for Socially Assistive Robot Tutors in In-Home Interventions

\author{Zhonghao Shi}
\email{zhonghas@usc.edu}
\orcid{https://orcid.org/0000-0002-2855-5863}
\affiliation{%
  \institution{University of Southern California}
  \city{Los Angeles}
  \state{USA}
  \postcode{90007}
}

\author{Thomas R Groechel}
\email{groechel@usc.edu}
\orcid{}
\affiliation{%
  \institution{University of Southern California}
  \city{Los Angeles}
  \state{USA}
  \postcode{90007}
}

\author{Shomik Jain}
\email{shomik.jain@alumni.usc.edu}
\orcid{https://orcid.org/0000-0001-5232-3264}
\affiliation{%
  \institution{University of Southern California}
  \city{Los Angeles}
  \state{USA}
  \postcode{90007}
}

\author{Kourtney Chima}
\email{kchima@usc.edu}
\orcid{}
\affiliation{
  \institution{University of Southern California}
  \city{Los Angeles}
  \state{USA}
  \postcode{90007}
}

\author{Ognjen (Oggi) Rudovic}
\email{orudovic@mit.edu}
\orcid{}
\affiliation{%
  \institution{Massachusetts Institute of Technology}
  \city{Cambridge}
  \state{USA}
  \postcode{02139}
}

\author{Maja J Matari\'c}
\email{mataric@usc.edu}
\orcid{https://orcid.org/0000-0001-8958-6666}
\affiliation{%
  \institution{University of Southern California}
  \city{Los Angeles}
  \state{USA}
  \postcode{90007}
}

\renewcommand{\shortauthors}{Shi et al.}

\begin{abstract}
Affect-aware socially assistive robotics (SAR) has shown great potential for augmenting interventions for children with autism spectrum disorders (ASD). However, current SAR cannot yet perceive the unique and diverse set of atypical cognitive-affective behaviors from children with ASD in an automatic and personalized fashion in long-term (multi-session) real-world interactions. To bridge this gap, this work designed and validated personalized models of arousal and valence for children with ASD using a multi-session in-home dataset of SAR interventions. By training machine learning (ML) algorithms with supervised domain adaptation (s-DA), the personalized models were able to trade off between the limited individual data and the more abundant less personal data pooled from other study participants. We evaluated the effects of personalization on a long-term multimodal dataset consisting of 4 children with ASD with a total of 19 sessions, and derived inter-rater reliability (IR) scores for binary arousal (IR = 83\%) and valence (IR = 81\%) labels between human annotators. Our results show that personalized Gradient Boosted Decision Trees (XGBoost) models with s-DA outperformed two non-personalized individualized and generic model baselines not only on the weighted average of all sessions, but also statistically ($p$ < .05) across individual sessions. This work paves the way for the development of personalized autonomous SAR systems tailored toward individuals with atypical cognitive-affective and socio-emotional needs.

\end{abstract}




\maketitle

\section{Introduction}
Within human-robot interaction (HRI), the field of socially assistive robotics (SAR) has emerged at the intersection of assistive robotics and socially interactive robotics~\cite{feil_defining}. Its central focus is to provide effective assistance and interventions through intelligent interactions with users~\cite{feil_defining, mataric_sar_handbook}, while also respecting socio-emotional needs of various users. As research in SAR is shifting from constrained laboratory settings toward real-world environments such as in-home interventions, it is critical for SAR systems to be able to perceive the user's emotional (i.e. affective) cues in order to provide naturalistic affect-aware interactions and interventions~\cite{breazeal2003emotion, picard2000affective}. Toward this end, recent research has shown that affect-aware SAR tutors have the potential to address the imbalance between cognitive and affective awareness found in contemporary robot-assisted teaching systems~\cite{spaulding2016affect}. However, existing SAR systems still lack the ability to elicit, perceive, and appropriately respond to user affect in a personalized fashion so that interventions can be tailored towards individual cognitive and affective needs along the learning process~\cite{woolf2009affect}. 

Both the promise and the challenges of SAR for personalized affect awareness are particularly amplified in the context of children with autism spectrum disorders (ASD).  ASD affects 1 in 64 in the United States and has a male-to-female ratio of 4:1~\cite{christensen2018prevalence}. Traditionally, therapists design interventions by using static toys as tools to induce open-ended and engaging interactions~\cite{kasari2018smarter}. More recently, research has shown success in adopting SAR tutors as a means of providing more effective interventions for children with ASD~\cite{scassellati2018improving}, due to their interactive and engaging nature as well as the robots' repetitive behaviors~\cite{scassellati2012robots}. However, every child with ASD has a unique profile of strengths, challenges, and autism-specific characteristics~\cite{stewart2009sensory}. The lack of personalized affect awareness hinders the ability of existing SAR tutors to perceive and respond to the unique and atypical cognitive-affective behaviors. Therefore, there is a need for affect-aware SAR tutors that can enable personalized SAR interventions for each child with ASD, with the goal of achieving positive cognitive and affective learning outcomes in the long-term.

% - motivations

Affect-aware SAR tutors need to be able to perceive a diverse set of cognitive-affective states in long-term interventions. For instance, engagement is an important metric for evaluating the effectiveness of SAR tutors. For this reason, most prior work has focused on applying supervised machine learning (ML) to enable robot perception of engagement directly from users' behavioral data (e.g., a child's vocalizations, facial and body expressions, and autonomic physiology data such as heart rate)~\cite{rudovic2018personalized, rudovic2018culturenet,jain2020modeling}. However, ideal SAR tutors still need to perceive a more diverse set of cognitive-affective states, such as confusion and frustration, in order to facilitate cognitive learning~\cite{kort2001affective}. Although the two-dimensional framework for describing arousal and valence has been extensively studied in HRI, computational models for arousal and valence for children with ASD have only been studied in single-session, laboratory settings~\cite{rudovic2018personalized}. Such settings pose many limitations; the feasibility of modeling children's valence and arousal needs further investigation in real-world conditions and over a longer period of time.

%  Therefore, this work introduces a novel long-term cognitive-affective HRI dataset with annotated arousal and valence labels from children with ASD.

Personalized affect-aware SAR for children with ASD needs to be designed and validated in long-term multi-session interventions in order to adequately capture and model children's emotional expressions to target stimuli (e.g., different learning activities as part of an intervention). Although prior work has focused on personalized robot perception for ASD, the methods are only trained and validated on single-session in-lab datasets. For example, \citet{javed2020towards} collected a dataset of five children with ASD in a single-session study, and evaluated/trained ML models in a child-specific manner (the models were trained/tested on the non-overlapping data of the same child). Such child-specific models suffer from the limited amount of data collected from each individual, as they are unable to leverage data from other participants during training. In another single-session study, \citet{rudovic2018personalized} proposed a personalized perception of affect deep networks (PPA-nets), where the model personalization was achieved by the tuning of the network layers to each child's culture, gender, and individual characteristics. However, it is difficult to generalize about the longer-term effects of the personalization since all of the data were collected in a single day-long session. Therefore, to better approximate longer-term use contexts, the impact of model personalization for affect awareness needs be evaluated on multi-session recordings of children's interactions with SAR. 

This paper proposes a novel approach to modeling personalization that is specifically tailored to multiple sessions of child-robot interactions of children with ASD in long-term in-home interventions. The sessions included in the dataset are a part of our month-long in-home deployments of SAR interventions with children with ASD. Specifically, the multi-session child data in this study were used to devise personalized ML models for the estimation of children's cognitive-affective states in terms of their valence and arousal levels, as described in Sec.~\ref{sec:affectawaresartutor}. The model personalization is attained by using the notion of supervised domain adaptation - an ML approach that leverages a small portion of supervised data to learn to adapt generic (population-level) models to each individual. The contributions of this work are summarized as follows: 
\begin{description}
    \item[$\bullet$ A novel multi-session real-world ASD dataset] This paper introduces a novel multi-session multimodal dataset of arousal and valence collected from long-term in-home interventions of SAR tutors with children with ASD. The dataset consists of four participants in 19 sessions; totalling 8 hours, 16 minutes and 20 seconds of intervention data. Arousal and valence were annotated as binary labels for each participant following standard definitions grounded in established practices from emotion science and automatic emotion recognition~\cite{paulmann2013valencearousal, gunes2010automatic}. The multimodal feature space consists of visual (body pose and facial), audio, and game performance features, extracted using OpenFace, OpenPose and Praat~\cite{cao2019openpose,Baltrusaitis2018OpenFace2F,boersma_praat}. To the best of our knowledge, this is the first dataset of its kind that enables research into the longitudinal modeling of valence-arousal behavioral cues of children with ASD using ML. More details can be found in Sec.~\ref{sec:participants}.
    
  \item[$\bullet$ Modeling Arousal/Valence in Long-Term Interventions:] This paper is the first to propose a design and validation of a computational models of arousal and valence from children with ASD on a multi-session real-world dataset of SAR interventions. Although prior work has analyzed arousal and valence in typically developing user populations where individual differences are less pronounced, no prior work to our knowledge has attempted to model valence and arousal from multi-session data of children with ASD. Specifically, we found that the generic XGBoost classifier (our baseline) outperformed the other model candidates (Feed-forward NNs, LogReg, SVM, and KNN) and achieved a 90\% AUROC score for arousal and 83\% AUROC score for valence across all sessions data. The results per child are found in Sec.~\ref{sec:alternativeapproaches}.
  
  \item[$\bullet$ Long-term Model Personalization:] By using the notion of the supervised domain adaptation (s-DA) based on loss-reweighting~\cite{chen2019crosstrainer}, we designed and validated the models for long-term model personalization from the data of multi-session interventions for children with ASD. We demonstrate that personalized models with s-DA significantly outperform their non-personalized counterparts (i.e., the individualized and generic models) in terms of the AUROC curve (with specific improvements over individualized models: arousal: +5\%, valence +4\%; and generic models: arousal: +2\%, valence +3\%). We show that these improvements are statistically significant across all four participants ($p$ < .05). We also show that the performance boost of personalized models was driven by an improved performance on the challenging negative class (low arousal/valence), while maintaining similar performance on the positive class (high arousal/valence). Results of the effects of model personalization are found in Sec.~\ref{sec:results}.
 
  \item[$\bullet$ Session-Based Model Evaluation:] Different from previous works that report the results in a traditional manner on single-session datasets, due to the nature of our data, we introduce a session-based model evaluation to capture more fine-grained performance of the models on our multi-session HRI dataset. Instead of adapting the percentage-based model evaluation methods used in past work~\cite{jain2020modeling}, we propose chronological train-test splits based on the recorded sessions; therefore, our method follows the same temporal dependence as real-world deployments of such a system. Consequently, this allows us to better quantify the effects of model personalization and its impact on performance in unseen sessions of the test participant. Results of session-based model evaluation method are found in Sec.~\ref{sec:modelevaluation}.

\end{description}

Our experimental evaluations show the importance of model personalization and gains in terms of valence-arousal estimation accuracy on multi-session data. Building upon our prior work on modeling engagement~\cite{jain2020modeling}, this is the first time that such long-term personalized model design and session-based evaluations have been performed on valence-arousal data of children with ASD across multiple-sessions and as part of in-home interventions. This has important implications for the design of future robot companions and tutors for children with ASD, paving the way for new personalized robot technologies for autism therapy and more effective long-term learning activities for all users.

% ---------------------------Related Work---------------------------------------

\section{Related Work}
 This section overviews background work relevant to the main contribution areas of this article: affect-aware SAR tutors (Sec.~\ref{sec:affectawaresartutor}), personalization for assisting ASD therapy (Sec.~\ref{sec:personalization}), and model evaluation for HRI (Sec.~\ref{sec:modelevaluation-back}).

% ---------------------------Personalization for ASD---------------------------------------

% ---------------------------Affect-Aware SAR Tutors---------------------------------------

\subsection{Affect-Aware SAR Tutors}
\label{sec:affectawaresartutor}

A significant body of SAR research has shown that autonomous SAR tutors enhance cognitive learning gains of children with ASD in a variety of settings~\cite{clabaugh2019escaping, ismail2019leveraging}. However, those tutoring systems are limited in their ability to autonomously perceive and respond to atypical affective behaviors of children with ASD~\cite{jain2020modeling}, hindering their ability to personalize toward the specific needs of each child~\cite{rudovic2018personalized, jain2020modeling, huang2014affect-awareness}. Previous research has studied perception of basic affective states (fear, anger, happiness, sadness, disgust, and surprise)~\cite{Celiktutan2018affect}, but child participants rarely experience fear or disgust during the learning process~\cite{d2007toward, gunes2019live}. In contrast, \citet{kort2001affective} addressed the interplay of affect and learning by modeling cognitive-affective states (engagement, confusion, frustration, etc.) that children with ASD naturally experience in the context of learning, using either a \textit{categorical} or \textit{dimensional} approach.

% ---------Affect-Aware SAR Tutors---------Categorical Approach---------------------------------------

{\it Categorical approaches} map affective states experienced during learning onto a set of basic cognitive-affective states such as engagement, confusion, frustration, boredom, or delight~\cite{d2011half}. Since user engagement is considered to be a particularly important metric for evaluating the effectiveness of SAR contexts, robot perception of user engagement has become a crucial capability of autonomous SAR systems~\cite{spaulding2016affect}. This is especially true for educational SAR tutors that must promote high user engagement levels in order to helps users achieve the desired cognitive learning gains~\cite{spaulding2016affect}. Consequently, the majority of prior work on developing affect-aware SAR tutors has focused on modeling user engagement~\cite{jain2020modeling}.

% --- paragraph of background on engagement modeling in general ---------
Most previous studies have trained and evaluated supervised ML classifiers on multimodal datasets for perception of engagement in HRI interactions. \citet{castellano2009detecting} and \citet{sanghvi2011automatic} trained supervised engagement classifiers with hand-crafted social, physiological, or task-based interaction features. To automate this manual feature engineering process, \citet{rudovic2019personalized} proposed a novel deep reinforcement learning architecture for estimation of engagement directly from raw video data, where high-dimensional features were automatically extracted using a pre-trained convolutional neural network (CNN), ResNet. However, due to the unique challenges of the atypical cognitive-affective behaviors of children with ASD, it is unclear how these models, trained and validated on a typically-developing population, would generalize to children with ASD.

% -------------------------------------------------------------

To further enable research on engagement of children with ASD, most previous studies collected multimodal datasets from single-session laboratory HRI studies. For instance, \citet{Rudovic2017} collected a multimodal dataset from single-session HRI interventions with children with ASD from different cultural background (Japan and Europe). Based on that dataset, \citet{rudovic2018personalized, rudovic2018culturenet} proposed a personalized deep learning framework to model engagement intensity of children with ASD that achieved an average agreement of 60\% with human experts while estimating user engagement on a continuous scale from -1 to 1. In another single-session study, \citet{javed2020towards} trained a CNN to model user social engagement. The proposed CNN-based model achieved the accuracy of 78\%, but it did not outperform the accuracy of tree-based random forest models (81\%), indicating that simpler models may work better in this setting.

Recently, \citet{jain2020modeling} demonstrated the feasibility of modeling user engagement in our more challenging long-term, in-home SAR deployment using the dataset also used in this work (but without model personalization; see Sec. 2.3 for more details on differences from our work). The authors found that Gradient Boosted Decision Trees (XGBoost) outperformed feedforward neural networks and achieved approximately 90\% accuracy for binary classification of user engagement~\cite{jain2020modeling}. However, despite this recent progress in modeling user engagement of children with ASD, engagement-aware SAR tutors are still unable to address the diverse cognitive-affective needs of children with ASD during the learning process, such as confusion and frustration that can be directly analyzed from the arousal/valence dimensions of affect that we tackle in this work. Although promoting improved engagement does help child participants to achieve higher cognitive learning gains~\cite{spaulding2016affect}, \citet{lepper1988socializing} showed that expert human tutors tend to devote the same amount of attention to child participants' emotional goals as they do to cognitive learning gains. Therefore, it is important for affect-aware SAR tutors to have the ability to perceive and respond to not just engagement but a diverse range of cognitive-affective states of children with ASD.

% However, the dataset included trivial cases of disengagement where participants had left the video frame. Since such disengaged frames can be detected with high confidence during the feature extraction process \cite{Baltrusaitis2018OpenFace2F,cao2019openpose}, that negatively impacts the model's performance on more challenging and important frames when the child is disengaged while still interacting with the SAR tutor.

% ---------Affect-Aware SAR Tutors---------Dimensional Approach---------------------------------------

\begin{figure}[h] 
    \centering
    \includegraphics[width=1\linewidth]{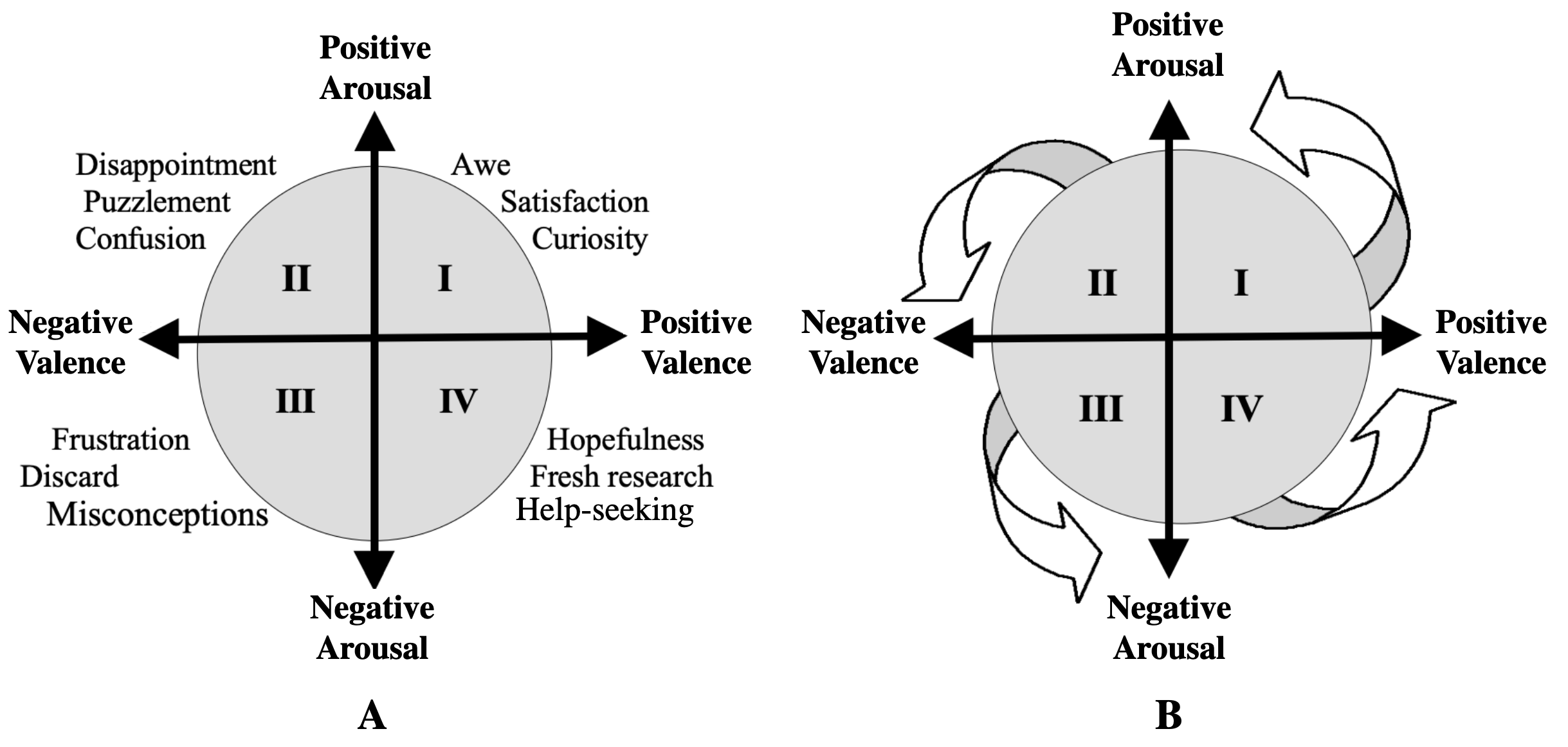}
    \caption{\textbf{A}: Based on \citet{kort2001affective}, two-dimensional arousal and valence framework for cognitive-affective states. States with positive valence (more pleasurable) are on the right; states with negative valence (more unpleasant) are on the left. Similarly, states with positive arousal (more constructive learning activity) are at the top; states with negative arousal (no learning activity) are at the bottom. \textbf{B}: Ideal circular flow of cognitive-affective states~\cite{kort2001affective}.} 
    \label{fig:interventions}
\end{figure}

{\it Dimensional approaches} map different cognitive-affective states onto a two-dimensional coordinate system consisting of perceived child participant valence and arousal, as shown in Figure~\ref{fig:interventions}~\cite{kort2001affective}. Furthermore, \citet{kort2001affective} categorized these cognitive-affective states into four major learning phases: 1) interested and curious about learning a new topic; 2) confused or puzzled by the learning topic; 3) frustrated or disengaged about some misconceptions; and 4) help-seeking for hints or fresh research. They suggested that autonomous tutoring systems should ideally first determine the child participant's cognitive-affective states, and then provide feedback based on that state to help the child participant be positively guided in the learning cycle in order to achieve both cognitive and emotional learning gains, as illustrated in Figure~\ref{fig:interventions}(B).

%- although many works on basic emotions, less work on CAS, especially for chidlre with ASD.
Although a large body of HRI research has studied emotion recognition using this two-dimensional framework~\cite{gunes2010automatic, coyne2020dimensional}, very few studies have designed computational models for arousal and valence from children with ASD~\cite{rudovic2018personalized, woolf2009affect}. \citet{rudovic2018personalized} modeled arousal and valence for children with ASD on a scale from -1 to 1 in a single-session in-lab study. However, due to the limitations of single-session studies in laboratory settings, the feasibility of modeling the children's valence and arousal requires further investigation in the real-world conditions and over a longer period of time. To this end, in this paper we address the  long-term, in-home SAR interventions for children with ASD. To our knowledge, this is the first time that the modeling of affective states of children with ASD has been attempted in real-world conditions. Toward that goal, we used a novel long-term dataset of arousal and valence collected from month-long in-home deployments of SAR tutors with children with ASD~\cite{clabaugh2018attentiveness}. Using this multi-session dataset, we demonstrate the feasibility of modeling arousal and valence of children with ASD in long-term in-home interventions, while showcasing the potential of using a two-dimensional approach to model a more diverse set of cognitive-affective states.

\subsection{Personalization in ASD}
\label{sec:personalization}

Personalization of the learning process is important for delivering effective educational interventions and is therefore a key feature of SAR tutors~\cite{greczek2014socially, Clabaugh2019frontiers}. This is particularly true in the ASD context since children with autism tend to have atypical and diverse ways of expressing their cognitive-affective states~\cite{stewart2009sensory}. To address this heterogeneity of the user population, past work has focused on studying the \textit{atypical behavioral patterns of children with ASD}, and \textit{personalized modeling} methods have been developed in the ASD context.

The work on HRI and SAR for autism is informed by the extensive research on ASD in developmental psychology and related fields~\cite{heidgerken2005survey, cabibihan2013robots}. It is well established that every child with ASD has a unique profile of strengths, challenges, and specific autism characteristics~\cite{stewart2009sensory}. This diversity makes it necessary to personalize SAR interventions to meet the unique needs of each child with ASD. Furthermore, past work found that children with ASD had a wide variety of reactions to robots and levels of ability to perform academic tasks~\cite{short2014variaty}, also highlighting the need for personalization.

Personalization has been studied extensively in SAR for ASD. While past work explored personalizing the robot action selection~\cite{leyzberg2018effect,Clabaugh2019frontiers,spaulding2019frustratingly}, this work focuses on studying the model personalization for robot perception of affect. \citet{javed2020towards} collected individual datasets for each child with ASD based on a single-session study. Separate models were trained and evaluated using each participant's dataset. However, this individual modeling approach did not leverage data from other participants for model training. To improve on this, in another single-session study, \citet{rudovic2018personalized} proposed a deep learning network, where personalization of the network was achieved using the demographic information (culture and gender), followed by individual network layers for each child. The personalized model outperformed the non-personalized baseline. However, due to the limitation of single-session study, the training and test sets were randomly sampled from the same session for each participant. Therefore, the observed improvement of personalized models is likely the result of the temporal dependence of instances between training and test sets. As suggested in the work, to eliminate the possible correlation between training and test sets, multi-session long-term studies need to be conducted, so the model can be evaluated on the next session to further validate the impact of model personalization.

Toward that end, we trained and evaluated our models on a multi-session dataset collected from month-long in-home deployments with four children with ASD. By applying supervised domain adaptation techniques for personalization, this paper shows that long-term personalized models outperform non-personalized baseline models. This paper further validates the need for long-term personalization of robot perception for children with ASD.

% ---------------------------Model Evaluation for HRI---------------------------------------

\subsection{Model Evaluation for HRI}
\label{sec:modelevaluation-back}
Due to the nature of our dataset, the design of the model evaluation scheme is critical for properly assessing the model's performance. Consequently, we first describe traditional evaluation protocols and then ours. Traditional methods for model evaluation differ in how the training and test sets are selected and how evaluation experiments are conducted. Model evaluation aims to estimate the accuracy induced by supervised ML algorithms so as to be able to determine the best-performing model during model selection~\cite{kohavi2001crossvalidation}. Work on affect-aware SAR tutors in ASD has primarily focused on two model evaluation methods: \textit{holdout} and \textit{cross-validation}.

{\it Holdout methods} partition the data into training and test sets based on a percentage parameter (e.g., 80\% training data to 20\% testing data). {\it Cross-validation methods} split the available data into $k$ mutually exclusive subsets. For each experiment, one of the $k$ subsets is used as the test set with the other $k-1$ subsets combined to form the training set. Similar to holdout methods, cross-validation methods commonly use random sampling to assign instances from the available data into each subset~\cite{kohavi2001crossvalidation}. 

% Holdout methods partition the data into training and test sets based on a percentage parameter (e.g., 80\% training data to 20\% testing data). With random sampling, instances of the available data are randomly assigned to either the training or test sets. To account for variance between each random sample, the holdout method is repeated $k$ times, and final results are derived as the average of all $k$ experiments \cite{kohavi2001crossvalidation}. Cross validation methods split the available data into $k$ mutually exclusive subsets. For each experiment, one of the $k$ subsets is used as the test set with the other $k-1$ subsets combined to form the training set. Similar to holdout methods, cross validation methods commonly use random sampling to assign instances from the available data into each subset. The reported results are also derived from the average of all $k$ experiments \cite{kohavi2001crossvalidation}. 

Recent work has used randomly sampled holdout or cross-validation for model evaluation. For instance, \citet{rudovic2018personalized} applied the randomly sampled holdout method to evaluate supervised ML models trained on a dataset collected from a single-session study involving children with ASD for user engagement and affect~\cite{rudovic2018personalized}. \citet{sanghvi2011automatic} and \citet{lala2017detection} performed randomly sampled stratified 10-fold cross-validation to evaluate the performance of engagement classifiers trained with different model candidates. More recently, \citet{javed2020towards} also used randomly sampled cross-validation for every subject's individual dataset to design individualized models for detecting engagement~\cite{javed2020towards}. 

However, due to the temporal nature of SAR deployments, random sampled model evaluation methods used in these past works are impractical for two major reasons. First, since SAR deployments obtain labeled training data chronologically before the testing data, model evaluation should also follow this temporal relationship to obtain accurate estimation of model performance. Random sampling violates this temporal relationship and possibly leads to having data from late in the interaction being assigned to the training set, and data from earlier in the interaction being assigned to the test set. Secondly, participant body pose and facial expressions are highly correlated for data instances that happen close in time. Therefore, with random sampling from the same single-session dataset, the training and test sets also become correlated, leading the model evaluation process to overestimate the model performance~\cite{jain2020modeling}.

To address the temporal nature of SAR datasets, \citet{jain2020modeling} applied a chronologically sampled holdout method for model evaluation, where an early subset of the data was used for training and the remaining subset was used for testing~\cite{jain2020modeling}. The work defined an {\it early subset} of the data as the first X\% of a user's data sorted chronologically. The evaluation experiments were conducted by varying the percentage of a data used for training from the first 10\% to the first 90\% of a user’s data, in increments of 10\%. The study successfully demonstrated the feasibility of recognizing user disengagement in real-world SAR settings. However, despite following the temporal relationship between data instances within each individual session, this method is still not practical for multi-session model evaluation. Since long-term SAR deployments obtain each session of deployments chronologically possibly on different days, the model can only be trained on the earlier sessions and tested on the next deployed session on another day. Since this percentage-based method is likely to split the first half of a session as the training set and the second half as test set, the model evaluation process violates the temporal dependence between sessions and leads to overestimation of models' performance. To address these limitations, {\it we introduce a session-based model evaluation method that follows the temporal dependence both across sessions and across data instances, so that the model evaluation process is able to estimate the real-world results of multi-session SAR deployments for children with ASD.}
More details on the proposed evaluation protocol can be found in Sec.~\ref{sec:modelevaluation}.
\section{Data Collection and Preprocessing}

\begin{figure}[t!] 
    \centering
    \includegraphics[width=0.9\linewidth]{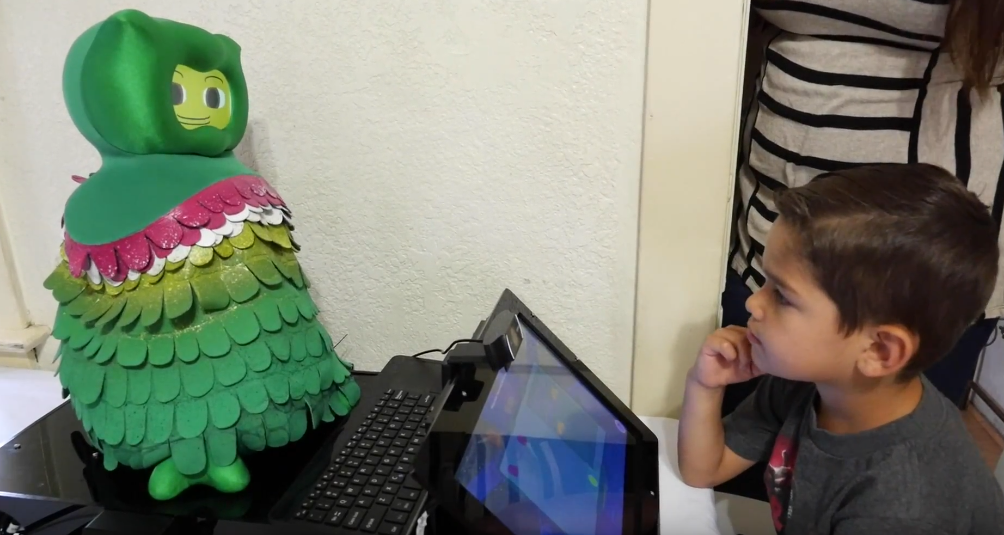}
    \caption{\textbf{System setup for the month-long in-home deployments.} Based on each child's cognitive performance, verbal and expressive feedback were provided by the SAR tutor to promote each child's social and math skill development. } 
    \label{fig:system}
\end{figure}
% - Study
The insights of this work result from a novel \textit{long-term dataset of arousal and valence data from children with ASD} collected in our month-long in-home deployments of SAR tutors~\cite{jain2020modeling}. In those deployments, each child participant with ASD interacted with a SAR tutor over many sessions by playing educational math games on a tablet, as shown in the Figure~\ref{fig:system}. Based on each child's cognitive performance, the SAR tutor provided both verbal and expressive feedback to promote the child's social and math skill development.

\subsection{Participants}
\label{sec:participants}
% - Data for each participant

The average age of the participants whose data were used for this work was 4.58$\pm$0.23 years old; 75\% were male and 25\% female, reflecting ASD rates in the recruited population. As discussed in our previous work~\cite{jain2020modeling}, the study was approved by the Institutional Review Board of the University of Southern California (UP-16-0075(v)). Informed consent forms were obtained from the  children's caregivers. The child participants had a clinical diagnoses of ASD in mild to moderate ranges according to the Diagnostic and Statistical Manual of Mental Disorders~\cite{american2013diagnostic}. The details about the SAR tutor and intervention design can be found in our previous publications~\cite{cait2019frontiers, pakkar2019designing}.

Due to the challenges of our month-long in-home data collections and the limitations of human annotation, four participants with sufficient and qualified data were selected to train and evaluate the personalized cognitive-affective models proposed in this study. Specifically, Participant 1 had five annotated sessions (total: 2hrs:8min:11sec); Participant 2 had six annotated sessions (total: 2hrs:8min:46sec); Participant 3 had four annotated sessions (total: 2hrs:16min:48sec); and Participant 4 had four annotated sessions (total: 1hr:42min:35sec). As discussed in our previous work~\cite{jain2020modeling}, data collected before and after the interventions were truncated for each session to only include the child-robot interaction between the first and last game in the dataset. Moreover, in this work, we also exclude trivial frames of disengagement where participants left the recorded view, so that the models can be trained/evaluated on the relevant portion of the data. 

\begin{figure}[t!] 
    \centering
    \includegraphics[width=1\linewidth]{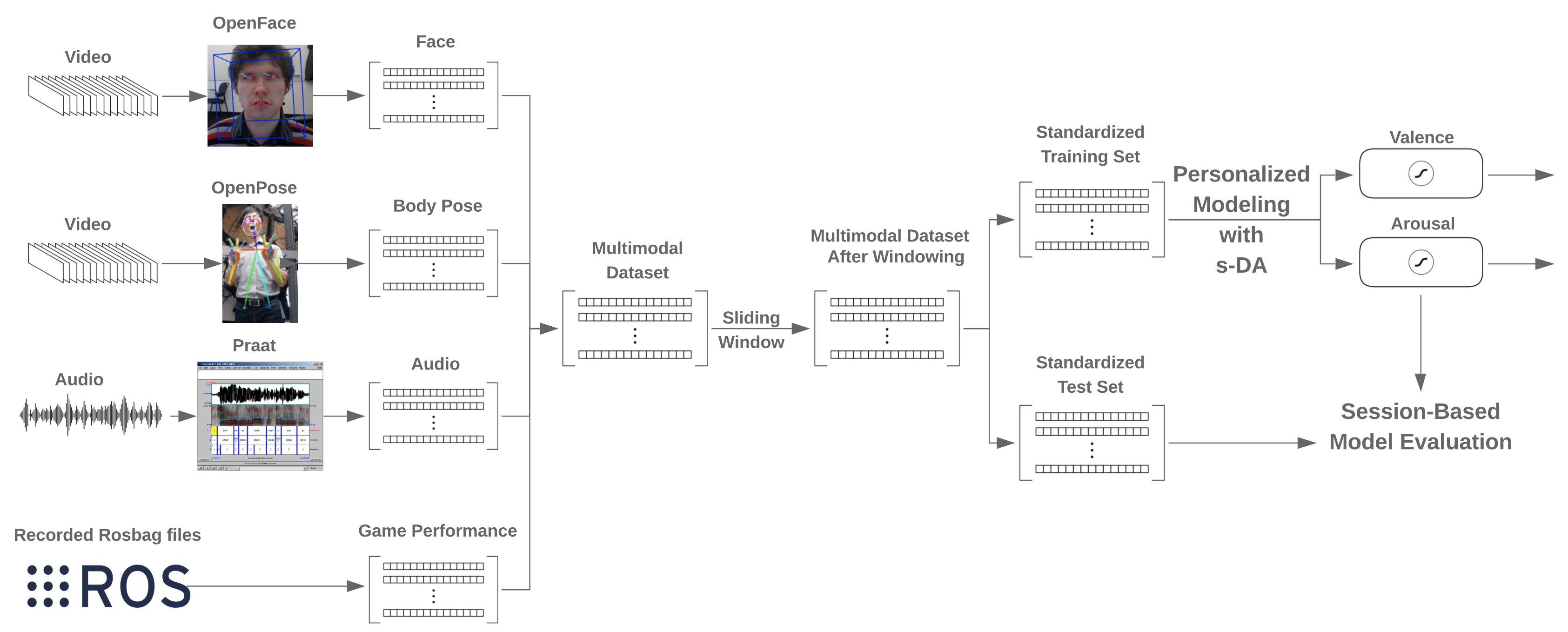}
    \caption{\textbf{Multimodal early fusion.} Multimodal features were extracted using open-source libraries (OpenFace~\cite{Baltrusaitis2018OpenFace2F}, OpenPose~\cite{cao2019openpose}, Praat~\cite{boersma_praat}, and ROS~\cite{quigley2009ros}) that could be used in a closed-loop systems, and then pre-processed using sliding window and standardization methods.} 
    \label{fig:multimodal}
\end{figure}

\subsection{Feature Extraction}
% - Feature Space
The feature space of our multimodal dataset consists of 115 visual (body pose and face) features, 6 audio features, and 16 game performance features. As discussed in our previous work~\cite{jain2020modeling}, the features were extracted using open-source libraries feasible for closed-loop use. The facial features extracted using OpenFace~\cite{Baltrusaitis2018OpenFace2F} included: a) eye gaze direction; b) facial expressions; c) face detection confidence value; and d) facial keypoint position. Praat~\cite{boersma_praat} was used to extract pitch, frequency, intensity, and harmonicity features from the recorded audio. Game performance features derived from the SAR system recording included: a) challenge level and type of game being played; b) number of games played in the session; c)  elapsed time in a session, game, and since the robot last talked; and d) number of incorrect responses. Additionally, in this work, we also used OpenPose~\cite{cao2019openpose} to extract body pose features including the number of people in the environment and keypoint positions for participant's body, feet, hands and face. 

\subsection{Definitions of Arousal and Valence}
% - Labels
To enable modeling of a diverse set of cognitive-affective user states, this work demonstrates the feasibility of modeling arousal and valence of children with ASD in long-term in-home interventions. Arousal and valence were annotated as binary labels (0 or 1) for each participant following standard definitions grounded in established practices from emotion science and automated emotion recognition~\cite{paulmann2013valencearousal, gunes2010automatic}. Specifically, in this study, the {\it valence label} refers to whether the child's emotion towards the SAR tutor is positive/neutral (1) or negative (0). The {\it arousal label} refers to whether the active level of the child's interaction with the SAR tutor is positive/neutral (1) or negative (0). The lead author of this work annotated the labels of arousal and valence for the 4 participants. To validate the absence of annotation bias, three additional annotators independently annotated 20~\% of the data for both valence and arousal labels. Inter-rater reliability scores were derived using Fleiss’ Kappa, and a reliability of 81\% and 83\% were achieved for valence and arousal, respectively, between the primary and verifying annotators.

\subsection{Data Preprocessing}
% - pre-processing

As shown in Figure~\ref{fig:multimodal}, we first performed early fusion of features from different modalities and pre-processed the dataset. After we concatenated features from different modalities, we obtained the raw multimodal dataset with a shape of $T \times M$, where $T$ is the total number of frames and $M$ is the total number of multimodal features. To address the uncertainty of extracted features for each individual frame, we applied a 3-second window with a 1-second shift to the raw multimodal dataset (i.e., 0 to 3 seconds, 1 to 4 seconds, 2 to 5 seconds, etc.). This 3-second window size was determined by the empirical observations of window size of participants' affective behaviors from our annotations. As discussed in our previous work~\cite{jain2020modeling}, within each window, the means and variances of continuous features were derived and added to the new dataset. For discrete features, a feature indicating whether features changed in the window was derived and added to the new dataset along with the means of the discrete features. This sliding window method also addressed the occasional frame dropping in OpenFace and OpenPose. Moreover, since the multimodal features were on different time scales, all data instances in both the training set and test set method were standardized to have zero mean and unit variance based on the means and variances of the training set~\cite{jain2020modeling}.

\subsection{Model Selection for Arousal/Valence}

We tested a collection of supervised ML model candidates during the model selection process. Past work has shown that tree-based XGBoost and Deep Neural Network (DNN) perform better than other model candidates for modeling user engagement~\cite{jain2020modeling,javed2020towards,rudovic2018personalized}. Consequently, we implemented XGBoost and DNN as the potential model candidates. Moreover, we also implemented Support Vector Machine (SVM) and K-Nearest Neighbor (KNN) as baseline models for comparison.  Because of large variances across sessions for each participant typical for ASD, hyperparameter tuning strategies such as grid search tend to overfit to the training data, so we used default hyperparameters for the model candidates. As discussed in our previous work~\cite{jain2020modeling}, we implemented the ML models using the open-source libraries shown as follows: Scikit-learn verion 0.22.1~\cite{pedregosa2011scikit}, XGBoost version 0.90~\cite{chen2016xgboost}, TensorFlow version 1.15.0~\cite{abadi2016tensorflow}, and Keras version 2.2.4~\cite{keras}. 

% Furthermore, to assess each modality’s importance for modeling arousal and valence, we trained and evaluated models trained on feature sets from different modalities--body pose, face, audio, and game performance--to evaluate the importance of features from each modality.

\section{Methods}
\label{sec:technical}
We introduce a \textit{personalized modeling method with supervised domain adaptation (s-DA)} for training personalized cognitive-affective models for children with ASD in long-term interventions. To validate the effects of long-term personalization, we introduce the \textit{session-based model evaluation method} to evaluate the personalized models by comparing them with two non-personalized baselines: individualized (participant-dependent) and generic (participant-independent) models. In this section, we describe these methods, describe the training and evaluation of the personalized models for arousal and valence of children with ASD, and further demonstrate the feasibility of modeling their cognitive-affective states in long-term in-home interventions.

\subsection{Personalization with Supervised Domain Adaptation}

As discussed in our previous work~\cite{jain2020modeling}, two major baseline methods have been used to train cognitive-affective models in long-term interventions: 1) individualized models trained only on available individual data; and 2) generic models trained only on available data from other participants. For individualized modeling, since participant-specific data are likely to be limited and imbalanced, individualized models are prone to overfitting. On the other hand, generic models are participant-independent, thus it is more challenging for such models to account for the heterogeneity in the data from the ASD population. We used both the generic and individualized models as baselines, and compared with our personalized modeling to explore the effects of the model personalization.

Our personalized modeling is achieved by using s-DA to obtain the optimal personalized cognitive-affective models for each child with ASD. By applying domain adaptation via {\it loss reweighting}~\cite{chen2019crosstrainer}, as described below, our method is able to trade off between the limited individual data and the more abundant but less personal data pooled from other participants, so that the best-performed personalized model can be obtained for each participant.

\begin{figure}[t!] 
    \centering
    \includegraphics[width=1.0\linewidth]{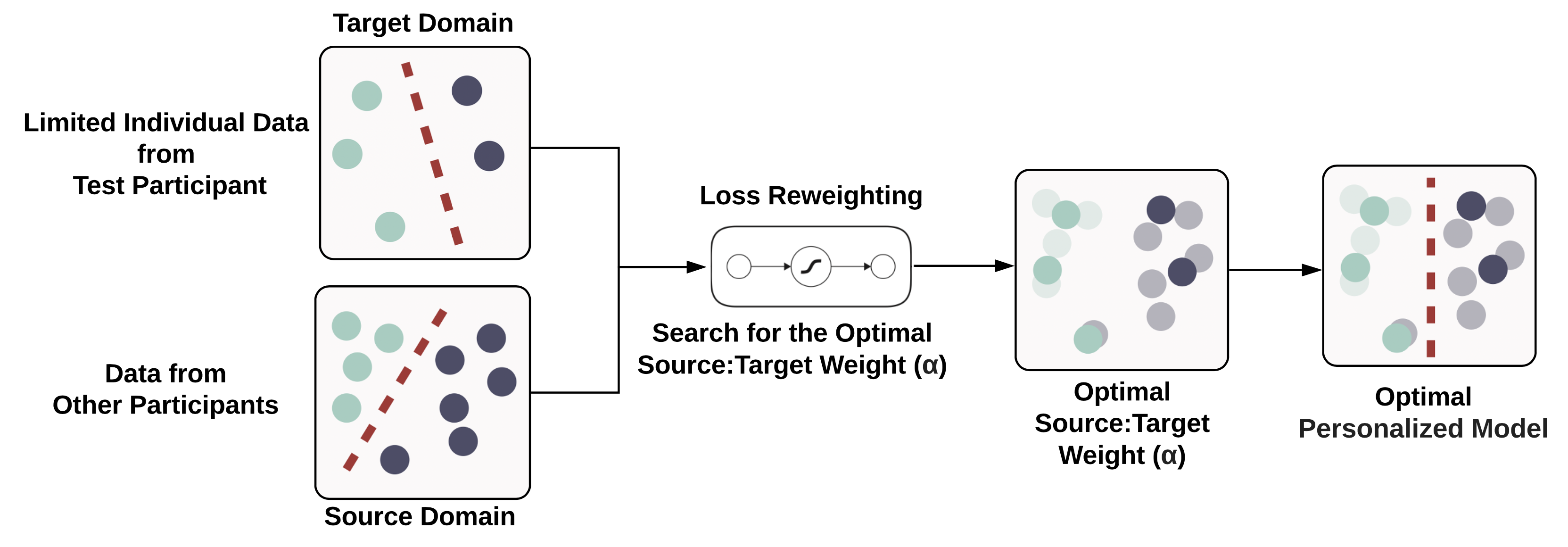}
    \caption{\textbf{Loss Reweighting.} Loss reweighting is an instance-based method for supervised domain adaptation that adjusts the loss function of a classifier to weight the relative importance of target and source datasets~\cite{chen2019crosstrainer}. It was applied in concert with personalized modeling to obtain the optimal personalized model for each child participant with ASD.}  
    \label{fig:reweighting}
\end{figure}

\begin{figure}[t!] 
    \centering
    \includegraphics[width=1\linewidth]{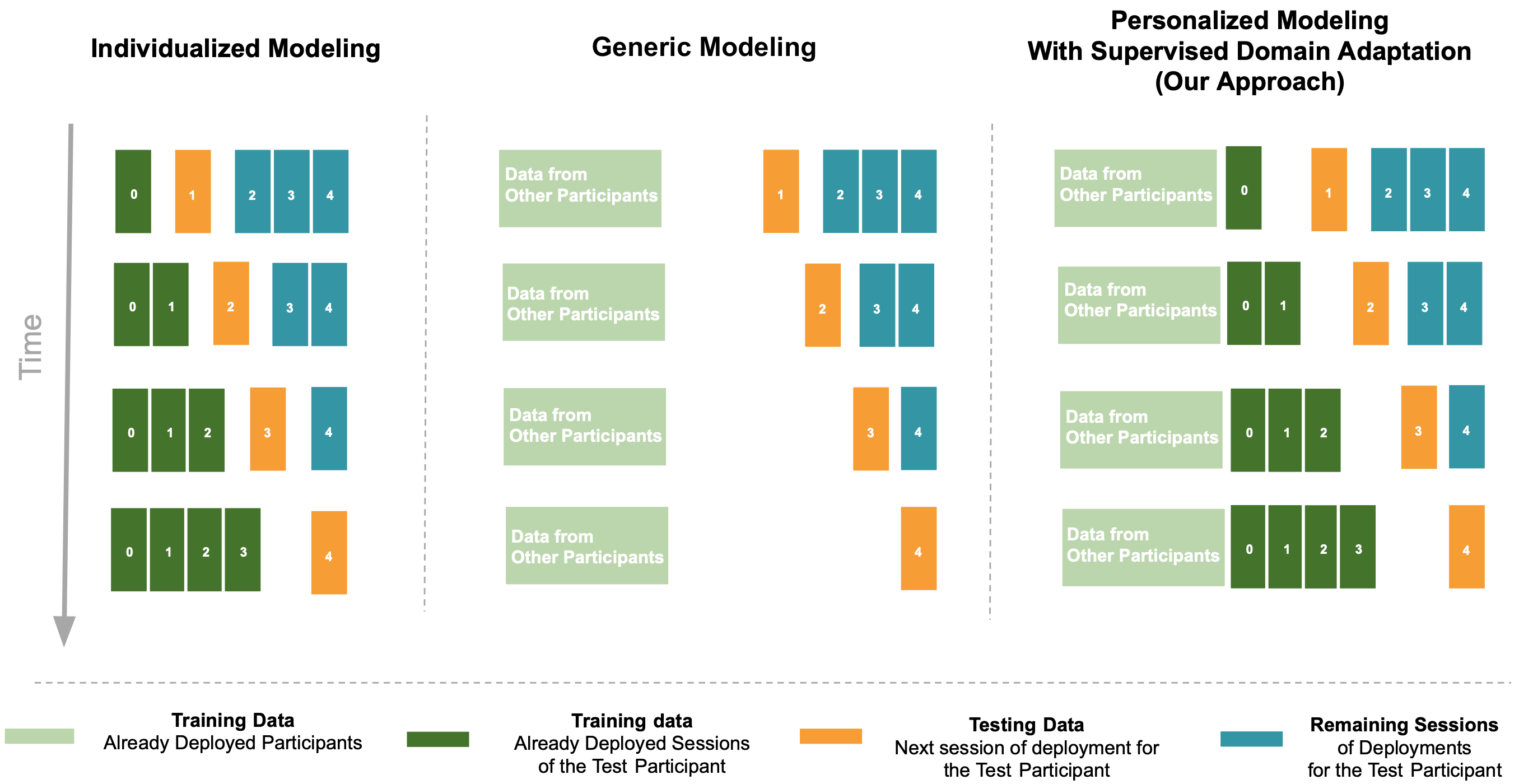}
    \caption{\textbf{A Comparison of Modeling Methods.} Left: individualized models are trained only on available individual data. Center: generic models are trained only on available data from other participants. Right: personalized models with s-DA are trained on available data from both the test data and other participants' data with the re-weighted loss function.} 
    \label{fig:modeling}
\end{figure}

Loss reweighting is an instance-based method for supervised domain adaptation that adjusts the loss function of a classifier to weight the relative importance of target and source datasets~\cite{chen2019crosstrainer}. In our personalized modeling method, shown in Figure~\ref{fig:reweighting}, data from other participants are used as the source domain, and the individual data from the test child participant with ASD are considered the target domain. By performing a hyper-parameter search with 5-fold cross-validation on the target individual data for the optimal weight $\alpha$ between the source and target domains, the loss reweighting s-DA technique searches for the optimal personalized model $h$ that minimizes the $\alpha$-error ${\epsilon}_{\alpha}(h)$. By adopting the optimized implementation of loss reweighting~\citet{chen2019crosstrainer}, this can formally be expressed as:

\begin{equation}
\label{equ:reweighting}
\hat{\epsilon}_{\alpha}(h)=\alpha \hat{\epsilon}_{T}(h)+(1-\alpha) \hat{\epsilon}_{S}(h), 
\end{equation}
where $\alpha$-error ${\epsilon}_{\alpha}(h)$ is a linear combination of the source domain (other participants' data) error $ \hat{\epsilon}_{T}(h)$ and the target domain (individual data) error $\hat{\epsilon}_{S}(h)$ for a given $\alpha \in[0,1]$. In the case of our multi-session data, we generalize the loss re-weighting function from Eq.~\ref{equ:reweighting} by accounting for its dependence on the sessions that accumulate over time as:

%here we need teh eq from above but in the context of session-based (k), and total num of sessions per child is K=1,\dots,N
\begin{equation}
\label{equ:reweighting2}
\hat{\epsilon}_{\alpha, k}(h)=\alpha \hat{\epsilon}_{T,k}(h)+(1-\alpha) \hat{\epsilon}_{S,k}(h), \text{where}\,\, k \in \{1,\dots,K\},
\end{equation}
where the session index $k$ is used to denote the data accumulated from all previous sessions up to session $k$. In this way, the personalized cognitive-affective models are gradually adapted to the target child's affective behaviors, while more individual data being included in a session-by-session fashion. Figure~\ref{fig:modeling} provides a comparison of our approach to other modeling methods.

\subsection{Session-Based Model Evaluation}
\label{sec:modelevaluation}

We introduce a session-based model evaluation to estimate the performance of the newly introduced models for multi-session data from SAR deployments for children with ASD. As discussed in Sec.~\ref{sec:modelevaluation}, randomly sampled and percentage-based model evaluation methods used in past work are not suitable for multi-session SAR deployments.  This work performs chronological train-test splits based on sessions; therefore, our method follows the same temporal dependencies both across sessions and across data instances as found in real-world deployments.

One round of session-based model evaluation for different modeling methods is illustrated in Figure~\ref{fig:modeling}. For each round, one out of the total $N$ participants is chosen to be the current test participant ($P$), and the data of the other $N-1$ participants are available for training. In each round, if there are $S$ sessions collected in the dataset for test participant $P$, $S-1$ experiments are then conducted chronologically for that test participant. For example, the current test participant illustrated in Figure~\ref{fig:modeling} has a total of 5 sessions, four experiments are conducted for that test participant. For each experiment, the train-test split is then based on $P$'s sessions and follows the same pattern of real-world deployments. For example, in the first experiment of individualized modeling, we assume that we have already obtained and annotated the first session of the test participant. This first session is used as the training set and the second session is used as the test set. In the next experiment of individualized modeling, the second session of the test participant is treated as if the deployment has completed and the data from the second session have been annotated. The model is then retrained on the union of the first and second sessions and tested on the third session. This process continues until every session except the first has been used as the test set. With $N$ different participants, $N$ rounds of experiments are conducted with different participants serving as the test participant for each round.

Similarly, with different training data, generic models and personalized models with s-DA can also be evaluated in the same fashion, as illustrated in Figure~\ref{fig:modeling}. Since all modeling methods are tested on the same set of sessions, the model evaluation can be kept consistent across different modeling methods. The final result is derived by averaging the models' performance in a weighted fashion on all tested sessions for each participant and comparing the performance between different modeling methods and model candidates. AUROC was used as the single-value metric to evaluate the overall performance of ML models on the binary classification of arousal and valence scores~\cite{bradley1997use}. Likewise, F1 scores were used to evaluate the models' performance on both the positive and negative classes of arousal and valence.

\section{Results}
 \label{sec:results}
\subsection{Effect of Model Personalization with s-DA}

We trained and evaluated the personalized models with s-DA with the aim of enabling automatic perception of arousal and valence of children with ASD during multi-session SAR interventions. As discussed in Sec.~\ref{sec:technical}, to obtain the optimal personalized models during training, the s-DA technique, used for our model personalization, enables the trade-off between the limited individual data and the more abundant data pooled from other participants. The effects of personalization were further validated by comparison with two baseline models: 1) individualized models trained only on available individual data; and 2) generic models trained only on data from other participants. 

As detailed in Sec.~\ref{sec:modelevaluation}, for each round of experiments, one of the four participants was chosen to be the current test participant, and data from the other three participants were considered to have been already obtained from deployments and available for training. Within each round, different modeling methods were trained and evaluated on the same sessions of the test participant using the session-based model evaluation method. 

\subsubsection{\textbf{Analysis of Participants.}}
\label{sec:weightedaverage}
Table~\ref{Tab:modelingmethods} reports models' AUROC scores of each experiment round with different test participant (P-1 to P-4) for both arousal and valence, so that we can compare the performance of the personalized models and the two baseline (individualized and generic) models. In addition, the weighted average AUROC scores of the four rounds of experiments for both arousal and valence are also reported to further examine the overall effects of the personalization. 

\begin{figure}[t!]
    \centering
    \includegraphics[width=1\linewidth]{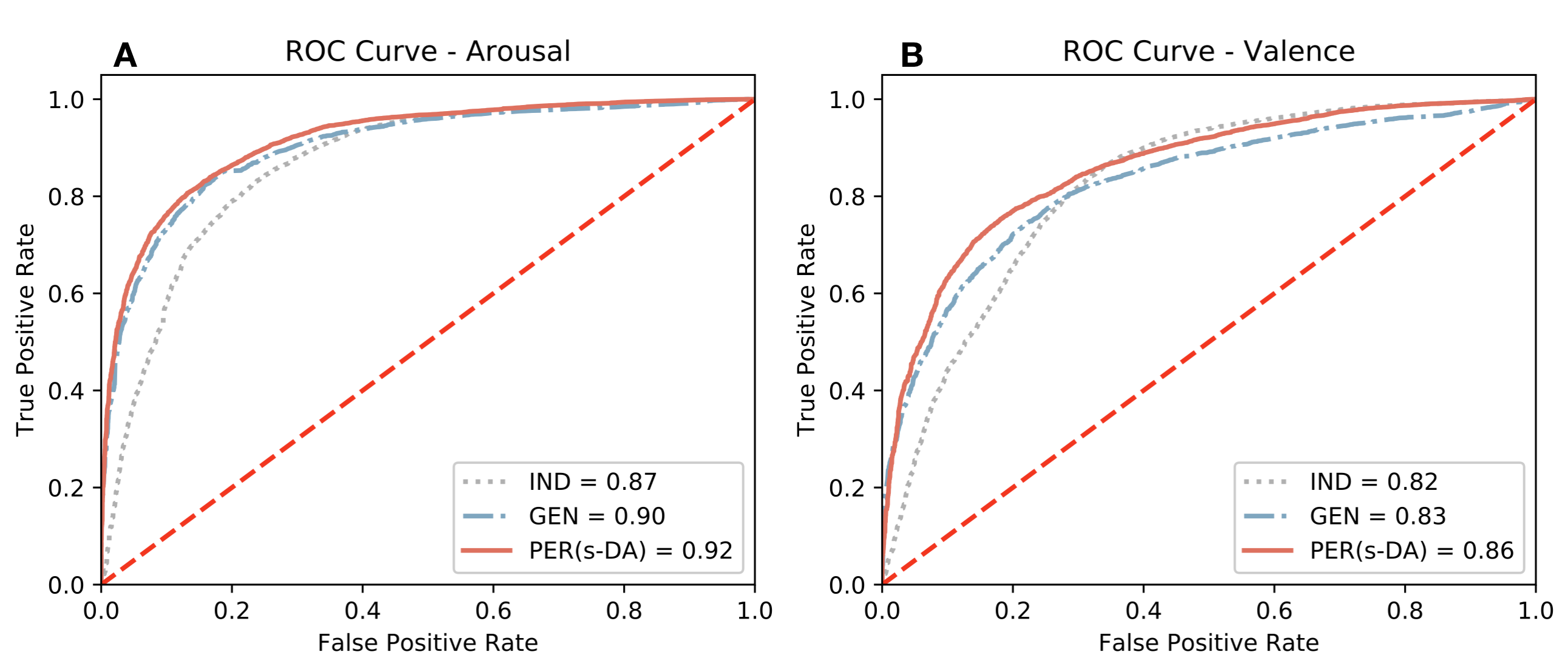}
    \caption{\textbf{ROC Curves of Personalized and Baseline XGBoost Models} Personalized (PER) models with s-DA outperformed both the individualized (IND) and generic (GEN) models using the metric of Area Under Receiver Operating Characteristic Curve (AUROC). } 
    \label{fig:comparison}
\end{figure}

Individually, personalized models outperformed the baseline models in six out of eight rounds of experiments across different test participants and tasks, and also performed at least as well as the baseline models in the other two rounds of experiments. As shown in Figure~\ref{fig:comparison}, overall, personalized models outperformed the best baseline models by approximately 2\% for arousal and 3\% for valence on a weighted average (wAVE) of the four test participants. More specifically, personalized models with s-DA achieved a 92\% AUROC score for arousal and an 86\% AUROC score for valence. The ROC curves show that our personalized models consistently outperform both model alternatives.

\begin{table}[t!]
\centering  

\begin{tabular}{?c|c|ccc?}
 \thickhline
Task & Test Participant-ID & IND & GEN & PER (s-DA) \\
 \thickhline
\multirow{5}{*}{ Arousal } & P-1  & 0.84 & \textbf{0.91} & \textbf{0.91} \\
&P-2  & 0.90 & 0.90 & \textbf{0.91} \\
&P-3  & 0.90 & 0.91 & \textbf{0.92} \\
&P-4  & 0.85 & 0.90 & \textbf{0.93} \\
\cline{2-5}
  & wAVE & 0.87 & 0.90 & \textbf{0.92} \\
 \thickhline
\multirow{5}{*}{ Valence } & P-1 & 0.79 & 0.82 & \textbf{0.84} \\
&P-2 & 0.85 & 0.86 & \textbf{0.88} \\
&P-3 & 0.84 & 0.83 & \textbf{0.87} \\
&P-4 & 0.81 & \textbf{0.87} & \textbf{0.87} \\
\cline{2-5}
 &wAVE  & 0.82 & 0.83 & \textbf{0.86} \\
 \thickhline
\end{tabular}

\caption{\textbf{Comparison of AUROC Scores of Personalized and Baseline XGBoost Models} Personalized (PER) models with s-DA outperformed both the individualized (IND) and generic (GEN) models on the weighted average (wAVE), and always performed at least as well as the baseline models across individual participants and tasks.}
\label{Tab:modelingmethods}
\end{table}

\begin{figure}[t!] 
    \centering
    \includegraphics[width=1\linewidth]{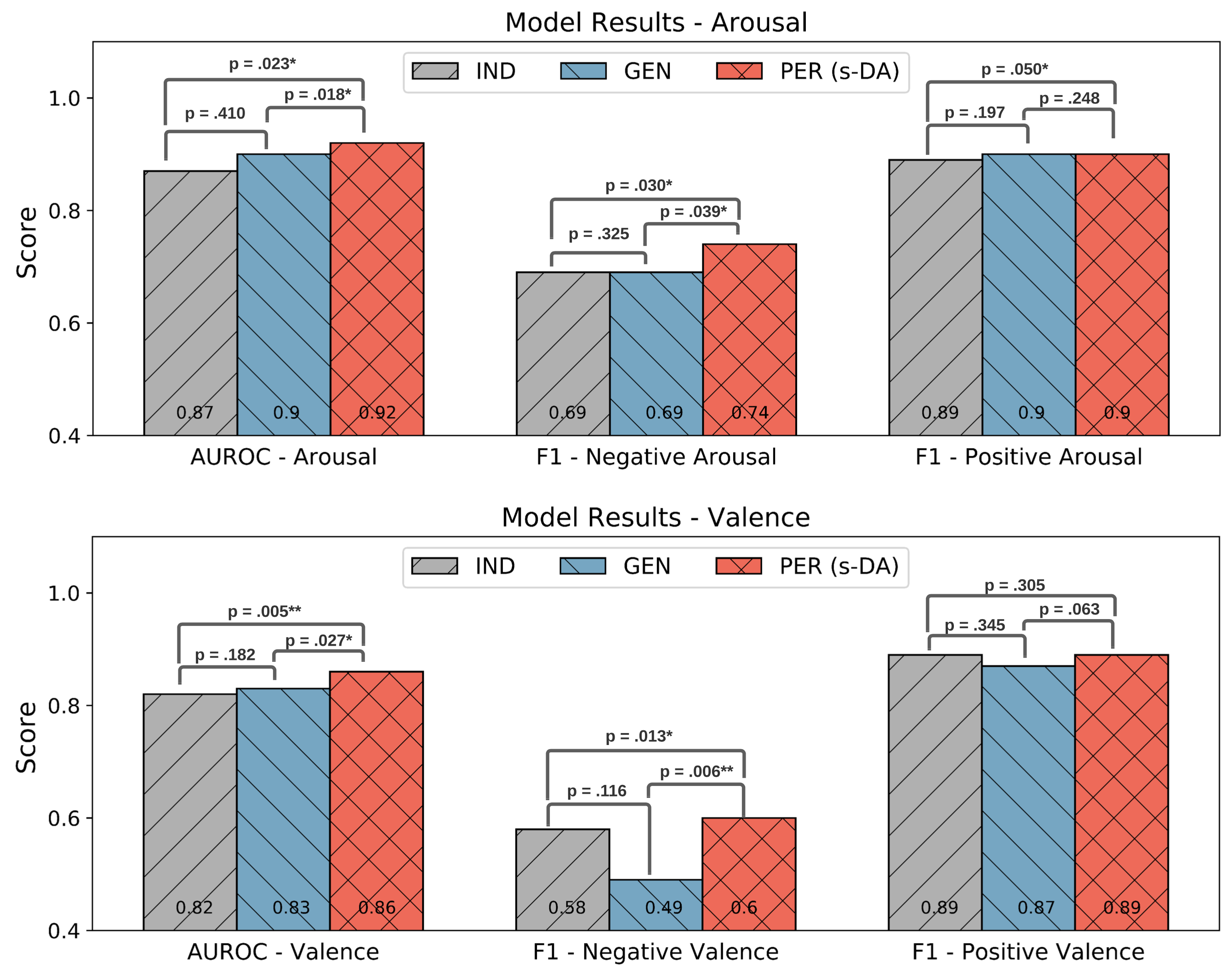}
    \caption{\textbf{Effect of Model Personalization with s-DA.} 1) Personalized (PER) models with s-DA significantly outperformed the baseline models in AUROC not only on the weighted average but also across individual tested sessions; 2) This performance boost of personalized models was driven by statistically higher performance for the negative class ($p$ < .05) across individual sessions, while maintaining performance on the positive class (no statistical difference).} 
    \label{fig:metrics}
\end{figure}

Moreover, as can be seen from Figure~\ref{fig:metrics}, personalized models especially outperformed baseline models on the negative class, while maintaining the same performance as the baseline models on the positive class (see also Sec. 6.1). For the negative class, personalized models achieved 2\% and 11\% improvements for valence over individualized and generic models, respectively, and also achieved 5\% improvements in F1 score for arousal over both baseline models. On the other hand, for the positive class, personalized models performed as the best baseline models on each task. Overall, personalized models achieved 90\% (arousal) and 89\% (valence) in F1 score for the positive instances, and 74 \% (arousal) and 60 \% (valence) in F1 score for the negative instances. This gap in performance between the two classes is due to the class imbalance against the negative class of arousal and valence. This is discussed in more details in Section~\ref{sec:classdistribution}.

\subsubsection{\textbf{Analysis of Sessions.}}
% Each session matters on its own. 
We also conducted one-sided Wilcoxon signed-rank tests to validate whether the effects of personalization for individual participants on the weighted average also extend to the individual sessions. As detailed in Figure~\ref{fig:metrics}, for both arousal and valence in AUROC, the one-sided Wilcoxon signed-rank tests validated that personalized models with s-DA also significantly ($p$ < .05) outperformed the two baseline models across sessions in AUROC, while there was no significant difference between individualized and generic models despite the difference in the weighted average. 

Furthermore, Wilcoxon signed-rank tests also validated that the performance boost of personalized models was driven by statistically higher performance for the negative class ($p$< .05) in F1 scores across individual sessions, while maintaining performance on the positive class (no statistical difference). More specifically, in AUROC, the Wilcoxon signed-rank tests found significantly higher AUROC for personalized over generic models (arousal: $Z$ = 2.101, $p$ = .018, $r$ = .384; valence: $Z$ = 1.931, $p$ = .027, $r$ = .353), and for personalized over individualized models (arousal: $Z$ = 1.988, $p$ = .023, $r$ = .363; valence: $Z$ = 2.556, $p$ = .005, $r$ = .467). However, there was no significant difference in AUROC between generic and individualized models (arousal: $Z$ = 0.227, $p$ = .41, $r$ = .041; valence: $Z$ = 0.909, $p$ = .166, $r$ = .182). For F1 scores of both negative and positive classes, the detailed results of Wilcoxon signed-rank tests can be found in the Appendix A.1.

\begin{table}[t!]
\centering  

\scalebox{0.58}{
\begin{tabular}{?c|c||ccccc|cccccc|cccc|cccc?}
% \multirow{2}{*}{Task } & \multirow{2}{*}{Model } & \multicolumn{15}{c}{ Test Participant }\\
% & &  \multicolumn{4}{c|}{ P-1 }  & \multicolumn{5}{c|}{ P-2 } & \multicolumn{3}{c|}{ P-3 }& \multicolumn{3}{c}{ P-4 }\\
\thickhline
\multirow{3}{*}{Task } & \multirow{3}{*}{Model } & \multicolumn{19}{c?}{ Test Participant }\\
\cline{3-21}
& &  \multicolumn{5}{c|}{ P-1 }  & \multicolumn{6}{c|}{ P-2 } & \multicolumn{4}{c|}{ P-3 }& \multicolumn{4}{c?}{ P-4 }\\
\cline{3-21}
& &  S-1&S-2&S-3&\multicolumn{1}{c|}{S-4}&\textbf{wAVE} & S-1&S-2&S-3&S-4 & \multicolumn{1}{c|}{ S-5 }&\textbf{wAVE}&S-1&S-2&\multicolumn{1}{c|}{S-3}&\textbf{wAVE}& S-1&S-2&\multicolumn{1}{c|}{S-3}&\textbf{wAVE}\\
 \thickhline
%  \cellcolor{blue!25}
 
\multirow{3}{*}{ Arousal }&IND        & 0.72 & 0.94 &  \cellcolor{blue!12}\textbf{0.95} & \multicolumn{1}{c|}{0.92} & 0.84& 0.83 &  \cellcolor{blue!12}\textbf{0.93} & 0.91 &  \cellcolor{blue!12}\textbf{0.93} & \multicolumn{1}{c|}{0.95} &0.90 &0.81 & 0.95 & \multicolumn{1}{c|}{ \cellcolor{blue!12}\textbf{0.97}}&0.90 & 0.74 &  \cellcolor{blue!12}\textbf{0.93} & \multicolumn{1}{c|}{ \cellcolor{blue!12}\textbf{0.93}} &0.85\\

&GEN        &  \cellcolor{blue!12}\textbf{0.90} & 0.94 & 0.94 & \multicolumn{1}{c|}{ \cellcolor{blue!12}\textbf{0.94}}& \cellcolor{blue!12}\textbf{0.91} &  \cellcolor{blue!12}\textbf{0.88} & 0.91 & 0.91 & 0.89 & \multicolumn{1}{c|}{0.94} &0.90& 0.85 & 0.95 & \multicolumn{1}{c|}{0.94}&0.91 & 0.89 & 0.88 & \multicolumn{1}{c|}{\cellcolor{blue!12}\textbf{0.93}} &0.90\\

&PER (s-DA) &  \cellcolor{blue!12}\textbf{0.90} &  \cellcolor{blue!12}\textbf{0.95} & 0.94 & \multicolumn{1}{c|}{0.93} & \cellcolor{blue!12}\textbf{0.91} & 0.85 &  \cellcolor{blue!12}\textbf{0.93} &  \cellcolor{blue!12}\textbf{0.93} &  \cellcolor{blue!12}\textbf{0.93} & \multicolumn{1}{c|}{ \cellcolor{blue!12}\textbf{0.96}} & \cellcolor{blue!12}\textbf{0.91}&  \cellcolor{blue!12}\textbf{0.86} &  \cellcolor{blue!12}\textbf{0.96} & \multicolumn{1}{c|}{0.94} &  \cellcolor{blue!12}\textbf{0.92}&  \cellcolor{blue!12}\textbf{0.91} &  \cellcolor{blue!12}\textbf{0.93} & \multicolumn{1}{c|}{ \cellcolor{blue!12}\textbf{0.93}}& \cellcolor{blue!12}\textbf{0.93} \\

\cline{1-21}

\multirow{3}{*}{ Valence }&IND & 0.60&0.84&\cellcolor{blue!12}\textbf{0.92}&\multicolumn{1}{c|}{0.81}&0.79&0.70&0.92&\cellcolor{blue!12}\textbf{0.89}&\cellcolor{blue!12}\textbf{0.95}&\multicolumn{1}{c|}{0.90}&0.85&0.74&\cellcolor{blue!12}\textbf{0.92}&\multicolumn{1}{c|}{\cellcolor{blue!12}\textbf{0.96}}&0.84&0.73&0.88&\multicolumn{1}{c|}{0.88} &0.81\\

&GEN & 0.67&\cellcolor{blue!12}\textbf{0.94}&0.77&\multicolumn{1}{c|}{\cellcolor{blue!12}\textbf{0.88}}&0.82&0.76&0.96&0.87&0.90&\multicolumn{1}{c|}{0.90}&0.86&0.73&0.89&\multicolumn{1}{c|}{0.93}&0.83&\cellcolor{blue!12}\textbf{0.86}&0.85&\multicolumn{1}{c|}{\cellcolor{blue!12}\textbf{0.93}} & \cellcolor{blue!12}\textbf{0.87}\\

&PER (s-DA) & \cellcolor{blue!12}\textbf{0.70}&0.91&0.90&\multicolumn{1}{c|}{0.86}&\cellcolor{blue!12}\textbf{0.84}&\textbf{0.80}&\cellcolor{blue!12}\textbf{0.97}&\cellcolor{blue!12}\textbf{0.89}&\cellcolor{blue!12}\textbf{0.95}&\multicolumn{1}{c|}{\cellcolor{blue!12}\textbf{0.91}}&\cellcolor{blue!12}\textbf{0.88}&\cellcolor{blue!12}\textbf{0.79}&0.91&\multicolumn{1}{c|}{\cellcolor{blue!12}\textbf{0.96}}&\cellcolor{blue!12}\textbf{0.87}&0.80&\cellcolor{blue!12}\textbf{0.90}&\multicolumn{1}{c|}{0.92}&\cellcolor{blue!12}\textbf{0.87} \\ 
\thickhline
\end{tabular}
}

\caption{\textbf{Session-by-Session Effects of Personalization With s-DA in AUROC.} Personalized (PER) models with s-DA were able to trade off between the alternating performance of individualized (IND) and generic (GEN) models, thus helping PER models with s-DA to outperform the baseline models not only on the weighted average but also statistically across sessions.}  
\label{tab:auroc}
\end{table}

In addition, for each individual session, personalized models with s-DA were able to trade off between the alternating performance of individualized and generic models, so it helped personalized models with s-DA to outperform the baseline models not only on the weighted average but also statistically across sessions. As can be noted from Table~\ref{tab:auroc}, the better-performed baseline model alternated between individualized and generic models, explaining the insignificant difference between individualized and generic models. As discussed in Sec.~\ref{sec:technical}, by taking advantage of individual data and data from other participants, personalized models achieved the best performance among the model candidates in the majority of test sessions. In the rest of the sessions, personalized models still consistently performed close to the better-performed baseline model. This further explains the significant increases from individualized/generic to personalized models across both participants and sessions. Finally, for the majority of the sessions/participants, the main increase in performance by the personalized models happens after adding the data of the first session of the target participant to the training set. After this, the performance trend remains relatively constant. This is the case for all but P-2, where we see a drop in the performance after the data from session 2 are added. This indicates that not all data from every session/participant are relevant for improving the model's performance in future sessions. One way to handle this is to learn how to select the most informative data and include only those in the adaptation set for model personalization. Various approaches have been proposed to tackle this issue by using the notion of Active/Reinforcement Learning (e.g., see~\cite{rudovic2019personalized}); this is outside of the scope of this work and will be explored in future work. 

\subsection{Class and Data Distribution}
\label{sec:classdistribution}

\begin{figure}[t!] 
    \centering
    \includegraphics[width=1\linewidth]{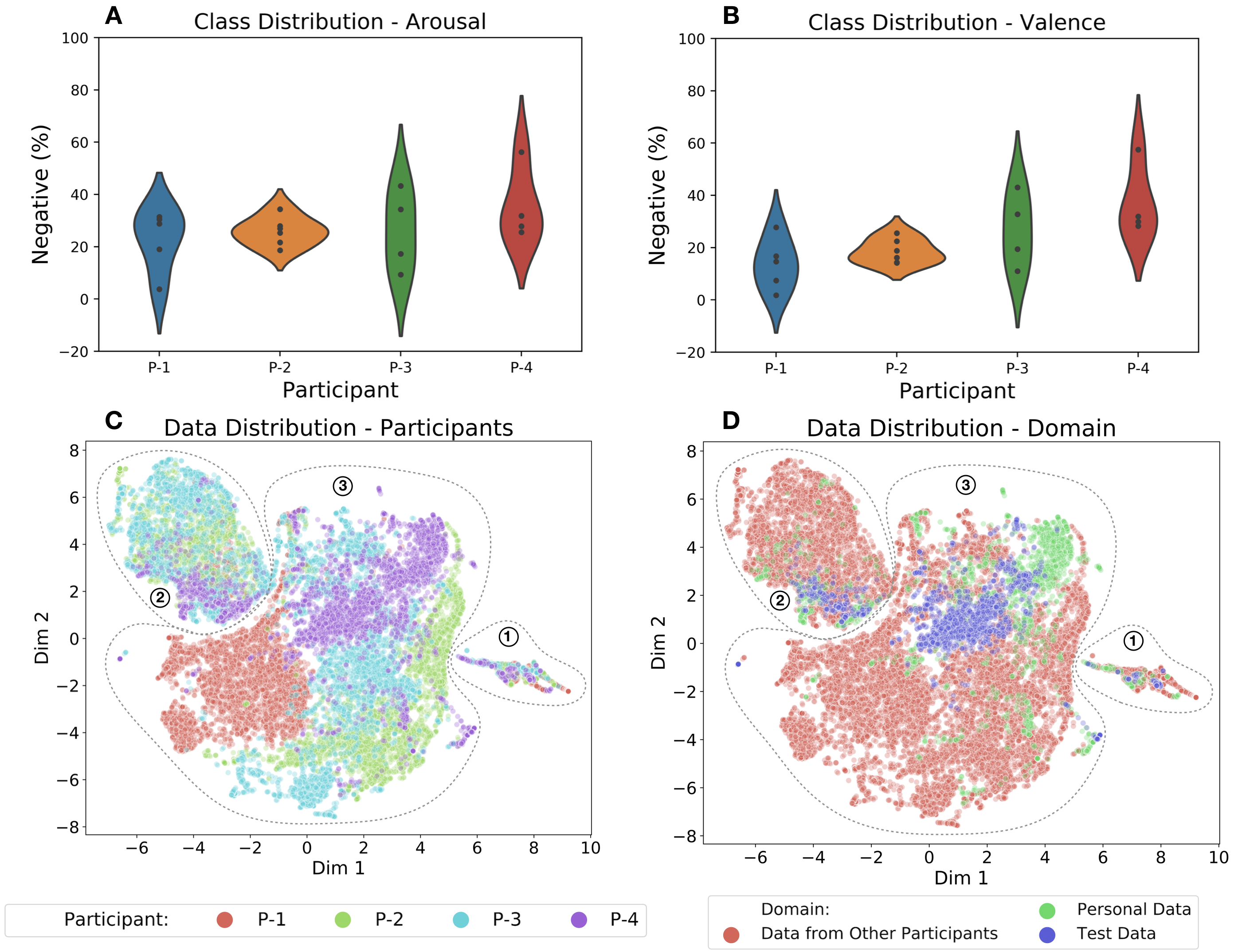}
    \caption{\textbf{Class and Data Distribution.} \textbf{A}: Percentage of negative arousal out of all arousal instances for each session of all participants; \textbf{B}: Percentage of negative valence out of all arousal instances for each session of all participants; \textbf{C}: Visualization of the data distribution for each participant using the T-distributed Stochastic Neighbor Embedding (t-SNE); \textbf{D}: Visualization of the data distribution for each domain of data within Participant 4 (Cluster 1: instances without OpenFace and OpenPose features; Cluster 2: instances without OpenFace features; Cluster 3: instances with complete set of features).} 
    \label{fig:distribution}
\end{figure}

As detailed in Figure~\ref{fig:distribution} (A, B), for both arousal and valence, we observed class imbalance in our multi-session dataset, helping to explain the less accurate performance of models on negative classes. 
Across different test participants with ASD, arousal and valence varied considerably from each other. From all instances of valence and arousal, average percentage of negative arousal for each child participant ranged from 22.7\% to 35.4\% with the standard deviation of 6\%, and average percentage of negative valence ranged from 14\% to 27\% with the standard deviation of 10\%. 

This variance between participants further supports the need for personalizing the models to each participant's unique cognitive-affective behaviors. Within each participant, different sessions also had substantial variance, highlighting the challenges of personalization in long-term in-home settings. The standard deviation between sessions of each participant ranged from 5\% to 14\% for arousal, and from 5\% to 15\% for valence.   

To better interpret the data distribution, this work also visualized the feature space of different participants using the T-distributed Stochastic Neighbor Embedding (t-SNE), a commonly used unsupervised dimensionality reduction technique.
As shown in Figure~\ref{fig:distribution}C, across different participants, high variance was observed as expected between the feature space of each participant. Within each participant, substantial variances were also observed in the feature space. For P-4, instead of clustering together, the data are located in smaller clusters separated from one another. This is consistent with our observations from annotation indicating that P-4 was more likely to move around during the interaction in the in-home environment compared to other participants.

Figure~\ref{fig:distribution} (C,D) shows that the extreme clusters (cluster 1 and 2) consist of instances where the participants are partially out of the recorded frames. More specifically, the cluster 2 consists of instances where OpenFace failed to detect the facial features and only part of the body pose features were extracted. For cluster 1 in Figure~\ref{fig:distribution}C, both OpenFace and OpenPose failed to detect the features from the recorded frames. The data instances in these two extreme clusters were mostly classified as negative arousal or valence. This is intuitive, since the models treated the missing features as negative behavioral cues of participants' arousal and valence. We could have removed these negative instances to make modeling less challenging, as we did when the child was completely missing from the camera-view.  Instead, we left those examples as they correspond to the partially observed data that still provide a proxy for the negative valence/arousal, since the participants are not showing positive valence/arousal toward the interaction with the robot.

% This supports THE EFFECT OF PERSONALIZATION

Moreover, we also visualized the feature space of different data domains (data from other participants, individual data, and test data). As shown in Figure~\ref{fig:distribution}D, in the main cluster (cluster 3), the distribution of the available individual data closely resembles the distribution of the test data. This further supports the importance of leveraging individual data and the value of personalization with s-DA.

\subsection{Alternative Modeling Approaches} 
\label{sec:alternativeapproaches}

\begin{table}[t!] 
\centering
\begin{tabular}{?c|c|ccccc?}
 \thickhline
 Test Participant & Task & KNN & SVM & LogReg & NN & XGBoost \\
 \thickhline
 \multirow{2}{*}{ wAVE }  &Arousal  & 0.76 & 0.86 & 0.86 & 0.84 & \textbf{0.90} \\
 \cline{2-7}
 & Valence & 0.70 & 0.74 & 0.79 & 0.78 & \textbf{0.83} \\
 \thickhline
\end{tabular}

\caption{\textbf{Results of Different Model Candidates} Tree-based XGBoost models were the most successful among the supervised ML model candidates we evaluated.}
\label{tab:modelselection}
\end{table}

From our initial model selection, we found that the tree-based XGBoost outperformed all the other model candidates when trained with the generic modeling method. In addition to its better performance, XGBoost also offers superior interpretability and requires less hyper-parameter tuning compared to the second best-performed NN. As shown in Table~\ref{tab:modelselection}, we also trained and tested the following model candidates with the generic modeling method: K-Nearest Neighbors (KNN), Support Vector Machines (SVM), Logistic Regression (LogReg), Neural Networks (NN), and Gradient Boosted Decision Trees (XGBoost). Due to the challenges of noisy and high-dimensional nature of the data we used, KNN performed poorly on both tasks (Arousal: drop of 14\%; Valence: drop of 13\%). LogReg was also not able to account for highly-nonlinear dependencies in the data (Arousal: drop of 4\%; Valence: drop of 4\%). On the other hand, the non-linear kernel method (SVM) also failed to achieve similar performance as XGBoost due to the absence of the hierarchical structure (Arousal: drop of 4\%; Valence: drop of 9\%). Finally, despite being hierarchical, NN had limited success because of the amount of data collected and the class imbalance (Arousal: drop of 6\%; Valence: drop of 5\%).

Moreover, we also explored the personalized modeling method with a unsupervised domain adaptation (u-DA) technique called CORrelation ALignment (CORAL)~\cite{sun2015return}. CORAL minimizes the domain shift (in our case, each child participant is seen as a different domain) by aligning the second-order statistics of source and target distributions, without requiring any target labels~\cite{sun2015return}. As detailed in Table~\ref{tab:unsupervised} in Appendix, the personalized models with u-DA performed poorly compared to the personalized models with s-DA (arousal: drop of 15\%; valence: drop of 15\%). This suggests that unsupervised domain adaptation methods such as CORAL may not be suitable for the noisy and nonstationary data typical for long-term interventions. Our future work will further explore this insight.

% \begin{table}[h] 

% \centering
% \begin{tabular}{?c|c|cc?}
%  \thickhline
%  Test Participant & Task &  All without Pose Features & All  \\
%  \thickhline
%  \multirow{2}{*}{ wAVE }  &Arousal  & 0.87 & \textbf{0.92} \\
%  \cline{2-4}
%  & Valence & 0.82 &  \textbf{0.86} \\
%  \thickhline

% \end{tabular}

% \caption{\textbf{Results of modality selection} }
% \label{tab:pose}
% \end{table}

\section{Discussion}

Personalized robot perception of arousal and valence from children with ASD is an important HRI research goal previously unexplored in the
context of long-term in-home SAR interventions. By introducing a novel multi-session dataset, this work is the first to train and evaluate personalized ML models of arousal and valence in this challenging context. This section focuses on discussing the major contributions and findings of this work in the following two areas: effect of personalization in long-term interventions (Sec.~\ref{sec:dis:effect}) and limitations and future work (Sec.~\ref{sec:dis:limit}). These contributions aim to inform the design of more affect-aware and personalized HRI, so that the effectiveness of SAR can be further improved to augment both the emotional and cognitive learning gains of children with ASD.

\subsection{Personalized Affect-Aware SAR in Long-Term Interventions}
\label{sec:dis:effect}
One of the main goals of this work was to demonstrate the feasibility of personalized cognitive-affective models in long-term HRI interventions for children with ASD. The results of this work suggest that supervised domain adaptation can effectively personalize ML models to better perceive the negative cognitive-affective behaviors of children with ASD. As shown in Sec.~\ref{sec:results}, personalized models with s-DA outperformed the best baseline models in AUROC on weighted average across each participant for both tasks. In particular, personalized models improved the predictions of negative arousal and valence when compared to the baseline models. Personalized models achieved 2\% and 11\% improvements in F1 score for negative valence over individualized and generic models, and also achieved 5\% improvements in F1 score for negative arousal over both the baseline models. This is particularly important because the purpose of our cognitive-affective models is to detect negative behaviors, so that corresponding feedback can be provided to help the child user be positively guided. Since the data imbalance and unique atypical negative affective behaviors from children with ASD further increased the challenges for perceiving negative affect, both baseline models performed worse on the negative class than the positive class. Therefore, personalized models with s-DA effectively improved the models' capability to perceive the unique negative arousal and valence of each child participant.

One-sided Wilcoxon signed-rank tests further validated that the better performance of personalization for individual participants on the weighted average also extended to the individual sessions. Statistically, personalized models with s-DA also significantly ($p$ < .05) outperformed the two baseline models across sessions in AUROC, while there was no significant difference between individualized and generic models despite the difference in the weighted average. This indicates that, with our dataset, it did not matter which method we chose between these two baselines; this may not hold for other datasets. However, for the dataset reported here, personalized models with s-DA were able to trade off between the strengths of both individualized and generic models.

The other main finding of this work is demonstrating the challenges of personalized affect awareness in a long-term real-world HRI setting due to the high variances and class imbalance. As detailed in Figure~\ref{fig:distribution}, we observed class imbalance in our long-term dataset, which caused the models to perform worse on the negative instances than the positive ones. This is expected since negative affective behaviors happened less frequently in our dataset than positive or neutral behaviors. Second, due to the unconstrained nature of the in-home setting, the more varied body movements and behaviors of the child participants resulted in more frequent failures of OpenFace or OpenPose capture of features. Therefore, the feature space was complex and challenging due to the substantial variances within each test participant. Due to the variances across sessions for each test participant, the variance between validation and test data further increase the challenges of long-term personalization.

Compared to the other alternative model candidates, XGBoost outperformed other supervised ML model candidates. This correlates with past findings for modeling engagement~\citep{jain2020modeling, javed2020towards} and expectations given the advantages of a heirarchical gradient boosted structure of XGBoost to perform on constrained data sets. Despite the recent success of deep learning, the challenging nature of the HRI study led to smaller amounts of noisy data with high class imbalance resulting in poorer NN performance. This work suggests that tree-based XGBoost is more suitable for real-world HRI dataset using constructed features extracted from libraries such as OpenFace and OpenPose. Furthermore, we also explored a naive instance-based unsupervised domain adaptation method called CORrelation ALignment (CORAL)~\cite{sun2015return}, and our results show that supervised domain adaptation had great advantages over CORAL. For this reason, we excluded those comparisons from the analysis reported here.

\subsection{Limitations and Future Work}
\label{sec:dis:limit}

The major limitations of this work come from the challenging nature of real-world HRI deployments for children with ASD in long-term interventions. The highly resource-intensive nature of conducting in-home deployments and collecting human annotations makes it difficult to obtain large-scale high-quality datasets of cognitive-affective behaviors from children with ASD. At the same time, the quality and scale of an annotated dataset is crucial for effectively training ML models such as NN-based approaches. In addition, the class imbalance of our dataset led to worse performance on negative arousal and valence. Despite the improvements from model personalization, the performance on negative affect still needs to be improved to enable effective interventions with robot feedback to help children users to be positively guided. Lastly, the high variance and noise in our autism dataset further challenges personalization. As discussed in Figure~\ref{fig:distribution}, the unconstrained in-home setting contributes to the high variance and noise in the dataset within each participant. Moreover, such data properties are expected from ASD datasets, since it is well known that every child with ASD has a unique profile of autism characteristics~\cite{stewart2009sensory}. Because both the source and target domain data in this work are collected from different children with ASD, this also potentially led to higher variance both within and across domains (and children participants).

One direction for future work is to explore annotation-efficient methods for model personalization so that the same modeling performance can be achieved more efficiently in terms of time an expense with fewer manual annotations. As we found in this work, not all the data from every session/participant are relevant for improving model performance in the future deployed sessions. This suggests that the amount of human annotations can be potentially reduced by selectively annotating the most informative personal data instances. Prior work has proposed to tackle this problem by applying Active/Reinforcement Learning~\cite{rudovic2019personalized}. This is outside of the scope of this work, but we plan to explore it in the future. 

Another promising approach is to continue using supervised domain adaptation with larger, more balanced and constrained datasets as source domains. For example, instead of using data from other children with ASD as the source domain, a more general dataset collected from typically developing children in a laboratory setting could potentially be used as the source domain. By keeping the target domain as the data from the target child with ASD, this method may allow personalized ML models to be trained more efficiently and effectively.

Finally, unsupervised domain adaptation (u-DA) remains a promising approach for personalized modeling with limited human annotation of the target domain data, despite our findings for CORAL. Preliminary work has shown that u-DA has great potential to successfully train personalized affective labels with no human annotation in the target domain~\cite{zen2014unsupervised}. However, most of the u-DA techniques have only been validated on unimodal (e.g., vision-only) datasets in constrained settings~\cite{wilson2020survey}. Consequently, open research questions remain as to how to effectively train personalized affective models with multimodal datasets involving children in unstructured settings like homes.

\subsection{Conclusions}
\label{sec:conclusion}

This work developed and validated personalized models for robot perception of arousal and valence from children with ASD in the previously unexplored context of a long-term in-home setting. Our motivation was to design personalized cognitive-affective models to perceive the unique cognitive-affective behaviors of each child with ASD. When deployed, these models should enable personalized interventions to help each child to be positively guided in the learning cycle and achieve improved cognitive and emotional learning gains.

We introduced a novel long-term multimodal dataset of arousal and valence collected from month-long in-home deployments of SAR tutors with children with ASD. Based on this dataset, we trained models with supervised domain adaptation (loss reweighting) to trade off between the limited individual data (target domain) and the more abundant generic data pooled from other participants (source domain). The models were also validated using a session-based method that follows the temporal dependence across sessions and data instances, so that the model evaluation process is able to estimate the real-world results of multi-session SAR deployments for children with ASD. The results showed that personalized models outperformed non-personalized (individualized and generic) models in both weighted average and statistically ($p$ < .05) across individual sessions. Personalized models especially improved the predictions of negative arousal and valence when compared to the baseline models. Our research demonstrates the potential of using supervised domain adaptation to improve the personalized affect awareness of SAR tutors with children with ASD in long-term interventions. We hope to inform the future development of personalized affect-aware robot tutors tailored towards individuals with atypical cognitive-affective needs such as children with ASD.

\section*{Acknowledgement}
This research was supported by the National Science Foundation Expedition in Computing Grant NSF IIS-1139148. The work of Ognjen Rudovic was funded by European Union H2020, Marie Curie Action - Individual Fellowship 2016-2019 (EngageMe 701236). The authors want to thank Balasubramanian Thiagarajan for his help with data analysis, and Sophia Pei, Ma$\ddot{y}$lis Whetsel, and Haocheng Yuan for their assistance with annotations. The authors are also very grateful to Caitlyn Clabaugh, Gisele Ragusa, Roxanna Pakkar, David Becerra, Eric C. Deng for their key roles in the original study design, recruitment, assessments, and more. The entire research team thanks the children and families who participated in the study that generated the dataset.

\newpage
\bibliographystyle{ACM-Reference-Format}
\bibliography{base}


\begin{thebibliography}{57}

%%% the Web address.

\ifx \showCODEN    \undefined \def \showCODEN     #1{\unskip}     \fi
\ifx \showDOI      \undefined \def \showDOI       #1{#1}\fi
\ifx \showISBNx    \undefined \def \showISBNx     #1{\unskip}     \fi
\ifx \showISBNxiii \undefined \def \showISBNxiii  #1{\unskip}     \fi
\ifx \showISSN     \undefined \def \showISSN      #1{\unskip}     \fi
\ifx \showLCCN     \undefined \def \showLCCN      #1{\unskip}     \fi
\ifx \shownote     \undefined \def \shownote      #1{#1}          \fi
\ifx \showarticletitle \undefined \def \showarticletitle #1{#1}   \fi
\ifx \showURL      \undefined \def \showURL       {\relax}        \fi
% invisible to TeX
\providecommand\bibfield[2]{#2}
\providecommand\bibinfo[2]{#2}
\providecommand\natexlab[1]{#1}
\providecommand\showeprint[2][]{arXiv:#2}

\bibitem[\protect\citeauthoryear{Abadi, Agarwal, Barham, Brevdo, Chen, Citro,
  Corrado, Davis, Dean, Devin, et~al\mbox{.}}{Abadi et~al\mbox{.}}{2016}]%
        {abadi2016tensorflow}
\bibfield{author}{\bibinfo{person}{Mart{\'\i}n Abadi}, \bibinfo{person}{Ashish
  Agarwal}, \bibinfo{person}{Paul Barham}, \bibinfo{person}{Eugene Brevdo},
  \bibinfo{person}{Zhifeng Chen}, \bibinfo{person}{Craig Citro},
  \bibinfo{person}{Greg~S Corrado}, \bibinfo{person}{Andy Davis},
  \bibinfo{person}{Jeffrey Dean}, \bibinfo{person}{Matthieu Devin},
  {et~al\mbox{.}}} \bibinfo{year}{2016}\natexlab{}.
\newblock \showarticletitle{Tensorflow: Large-scale machine learning on
  heterogeneous distributed systems}.
\newblock \bibinfo{journal}{\emph{arXiv preprint arXiv:1603.04467}}
  (\bibinfo{year}{2016}).
\newblock


\bibitem[\protect\citeauthoryear{Association et~al\mbox{.}}{Association
  et~al\mbox{.}}{2013}]%
        {american2013diagnostic}
\bibfield{author}{\bibinfo{person}{American~Psychiatric Association}
  {et~al\mbox{.}}} \bibinfo{year}{2013}\natexlab{}.
\newblock \bibinfo{booktitle}{\emph{Diagnostic and statistical manual of mental
  disorders (DSM-5{\textregistered})}}.
\newblock \bibinfo{publisher}{American Psychiatric Pub}.
\newblock


\bibitem[\protect\citeauthoryear{Baltrusaitis, Zadeh, Lim, and
  Morency}{Baltrusaitis et~al\mbox{.}}{2018}]%
        {Baltrusaitis2018OpenFace2F}
\bibfield{author}{\bibinfo{person}{Tadas Baltrusaitis}, \bibinfo{person}{Amir
  Zadeh}, \bibinfo{person}{Yao~Chong Lim}, {and}
  \bibinfo{person}{Louis-Philippe Morency}.} \bibinfo{year}{2018}\natexlab{}.
\newblock \showarticletitle{OpenFace 2.0: Facial Behavior Analysis Toolkit}.
\newblock \bibinfo{journal}{\emph{2018 13th IEEE International Conference on
  Automatic Face and Gesture Recognition (FG 2018)}} (\bibinfo{year}{2018}),
  \bibinfo{pages}{59--66}.
\newblock


\bibitem[\protect\citeauthoryear{Boersma}{Boersma}{2002}]%
        {boersma_praat}
\bibfield{author}{\bibinfo{person}{Paul Boersma}.}
  \bibinfo{year}{2002}\natexlab{}.
\newblock \bibinfo{title}{Praat, a system for doing phonetics by computer}.
\newblock
\newblock


\bibitem[\protect\citeauthoryear{Bradley}{Bradley}{1997}]%
        {bradley1997use}
\bibfield{author}{\bibinfo{person}{Andrew~P Bradley}.}
  \bibinfo{year}{1997}\natexlab{}.
\newblock \showarticletitle{The use of the area under the ROC curve in the
  evaluation of machine learning algorithms}.
\newblock \bibinfo{journal}{\emph{Pattern recognition}} \bibinfo{volume}{30},
  \bibinfo{number}{7} (\bibinfo{year}{1997}), \bibinfo{pages}{1145--1159}.
\newblock


\bibitem[\protect\citeauthoryear{Breazeal}{Breazeal}{2003}]%
        {breazeal2003emotion}
\bibfield{author}{\bibinfo{person}{Cynthia Breazeal}.}
  \bibinfo{year}{2003}\natexlab{}.
\newblock \showarticletitle{Emotion and sociable humanoid robots}.
\newblock \bibinfo{journal}{\emph{International journal of human-computer
  studies}} \bibinfo{volume}{59}, \bibinfo{number}{1-2} (\bibinfo{year}{2003}),
  \bibinfo{pages}{119--155}.
\newblock


\bibitem[\protect\citeauthoryear{Cabibihan, Javed, Ang, and Aljunied}{Cabibihan
  et~al\mbox{.}}{2013}]%
        {cabibihan2013robots}
\bibfield{author}{\bibinfo{person}{John-John Cabibihan}, \bibinfo{person}{Hifza
  Javed}, \bibinfo{person}{Marcelo Ang}, {and} \bibinfo{person}{Sharifah~Mariam
  Aljunied}.} \bibinfo{year}{2013}\natexlab{}.
\newblock \showarticletitle{Why robots? A survey on the roles and benefits of
  social robots in the therapy of children with autism}.
\newblock \bibinfo{journal}{\emph{International journal of social robotics}}
  \bibinfo{volume}{5}, \bibinfo{number}{4} (\bibinfo{year}{2013}),
  \bibinfo{pages}{593--618}.
\newblock


\bibitem[\protect\citeauthoryear{{Cao}, {Hidalgo Martinez}, {Simon}, {Wei}, and
  {Sheikh}}{{Cao} et~al\mbox{.}}{2019}]%
        {cao2019openpose}
\bibfield{author}{\bibinfo{person}{Z. {Cao}}, \bibinfo{person}{G. {Hidalgo
  Martinez}}, \bibinfo{person}{T. {Simon}}, \bibinfo{person}{S. {Wei}}, {and}
  \bibinfo{person}{Y.~A. {Sheikh}}.} \bibinfo{year}{2019}\natexlab{}.
\newblock \showarticletitle{OpenPose: Realtime Multi-Person 2D Pose Estimation
  using Part Affinity Fields}.
\newblock \bibinfo{journal}{\emph{IEEE Transactions on Pattern Analysis and
  Machine Intelligence}} (\bibinfo{year}{2019}).
\newblock


\bibitem[\protect\citeauthoryear{Castellano, Pereira, Leite, Paiva, and
  McOwan}{Castellano et~al\mbox{.}}{2009}]%
        {castellano2009detecting}
\bibfield{author}{\bibinfo{person}{Ginevra Castellano},
  \bibinfo{person}{Andr{\'e} Pereira}, \bibinfo{person}{Iolanda Leite},
  \bibinfo{person}{Ana Paiva}, {and} \bibinfo{person}{Peter~W McOwan}.}
  \bibinfo{year}{2009}\natexlab{}.
\newblock \showarticletitle{Detecting user engagement with a robot companion
  using task and social interaction-based features}. In
  \bibinfo{booktitle}{\emph{Proceedings of the 2009 international conference on
  Multimodal interfaces}}. \bibinfo{pages}{119--126}.
\newblock


\bibitem[\protect\citeauthoryear{Celiktutan, Sariyanidi, and Gunes}{Celiktutan
  et~al\mbox{.}}{2018}]%
        {Celiktutan2018affect}
\bibfield{author}{\bibinfo{person}{Oya Celiktutan}, \bibinfo{person}{Evangelos
  Sariyanidi}, {and} \bibinfo{person}{Hatice Gunes}.}
  \bibinfo{year}{2018}\natexlab{}.
\newblock \showarticletitle{Computational Analysis of Affect, Personality, and
  Engagement in Human--Robot Interactions}.
\newblock In \bibinfo{booktitle}{\emph{Computer Vision for Assistive
  Healthcare}}. \bibinfo{publisher}{Elsevier}, \bibinfo{pages}{283--318}.
\newblock


\bibitem[\protect\citeauthoryear{Chen, Gan, Rong, Suri, and Bailis}{Chen
  et~al\mbox{.}}{2019}]%
        {chen2019crosstrainer}
\bibfield{author}{\bibinfo{person}{Justin Chen}, \bibinfo{person}{Edward Gan},
  \bibinfo{person}{Kexin Rong}, \bibinfo{person}{Sahaana Suri}, {and}
  \bibinfo{person}{Peter Bailis}.} \bibinfo{year}{2019}\natexlab{}.
\newblock \showarticletitle{CrossTrainer: Practical Domain Adaptation with Loss
  Reweighting}. In \bibinfo{booktitle}{\emph{Proceedings of the 3rd
  International Workshop on Data Management for End-to-End Machine Learning}}.
  \bibinfo{pages}{1--10}.
\newblock


\bibitem[\protect\citeauthoryear{Chen and Guestrin}{Chen and Guestrin}{2016}]%
        {chen2016xgboost}
\bibfield{author}{\bibinfo{person}{Tianqi Chen} {and} \bibinfo{person}{Carlos
  Guestrin}.} \bibinfo{year}{2016}\natexlab{}.
\newblock \showarticletitle{Xgboost: A scalable tree boosting system}. In
  \bibinfo{booktitle}{\emph{Proceedings of the 22nd acm sigkdd international
  conference on knowledge discovery and data mining}}.
  \bibinfo{pages}{785--794}.
\newblock


\bibitem[\protect\citeauthoryear{Chollet}{Chollet}{2015}]%
        {keras}
\bibfield{author}{\bibinfo{person}{Francois Chollet}.}
  \bibinfo{year}{2015}\natexlab{}.
\newblock \bibinfo{title}{Keras}.
\newblock \bibinfo{howpublished}{https://keras.io}.
\newblock


\bibitem[\protect\citeauthoryear{Christensen, Braun, Baio, Bilder, Charles,
  Constantino, Daniels, Durkin, Fitzgerald, Kurzius-Spencer,
  et~al\mbox{.}}{Christensen et~al\mbox{.}}{2018}]%
        {christensen2018prevalence}
\bibfield{author}{\bibinfo{person}{Deborah~L Christensen}, \bibinfo{person}{Kim
  Van~Naarden Braun}, \bibinfo{person}{Jon Baio}, \bibinfo{person}{Deborah
  Bilder}, \bibinfo{person}{Jane Charles}, \bibinfo{person}{John~N
  Constantino}, \bibinfo{person}{Julie Daniels}, \bibinfo{person}{Maureen~S
  Durkin}, \bibinfo{person}{Robert~T Fitzgerald}, \bibinfo{person}{Margaret
  Kurzius-Spencer}, {et~al\mbox{.}}} \bibinfo{year}{2018}\natexlab{}.
\newblock \showarticletitle{Prevalence and characteristics of autism spectrum
  disorder among children aged 8 years—autism and developmental disabilities
  monitoring network, 11 sites, United States, 2012}.
\newblock \bibinfo{journal}{\emph{MMWR Surveillance Summaries}}
  \bibinfo{volume}{65}, \bibinfo{number}{13} (\bibinfo{year}{2018}),
  \bibinfo{pages}{1}.
\newblock


\bibitem[\protect\citeauthoryear{Clabaugh, Jain, Thiagarajan, Shi, Mathur,
  Mahajan, et~al\mbox{.}}{Clabaugh et~al\mbox{.}}{2018}]%
        {clabaugh2018attentiveness}
\bibfield{author}{\bibinfo{person}{Caitlyn Clabaugh}, \bibinfo{person}{Shomik
  Jain}, \bibinfo{person}{Balasubramanian Thiagarajan},
  \bibinfo{person}{Zhonghao Shi}, \bibinfo{person}{Leena Mathur},
  \bibinfo{person}{K Mahajan}, {et~al\mbox{.}}}
  \bibinfo{year}{2018}\natexlab{}.
\newblock \showarticletitle{Attentiveness of children with diverse needs to a
  socially assistive robot in the home}. In \bibinfo{booktitle}{\emph{2018
  International Symposium on Experimental Robotics}}.
\newblock


\bibitem[\protect\citeauthoryear{Clabaugh, Mahajan, Jain, Pakkar, Becerra, Shi,
  Deng, Lee, Ragusa, and Matari{\'c}}{Clabaugh et~al\mbox{.}}{2019b}]%
        {Clabaugh2019frontiers}
\bibfield{author}{\bibinfo{person}{Caitlyn Clabaugh}, \bibinfo{person}{Kartik
  Mahajan}, \bibinfo{person}{Shomik Jain}, \bibinfo{person}{Roxanna Pakkar},
  \bibinfo{person}{David Becerra}, \bibinfo{person}{Zhonghao Shi},
  \bibinfo{person}{Eric Deng}, \bibinfo{person}{Rhianna Lee},
  \bibinfo{person}{Gisele Ragusa}, {and} \bibinfo{person}{Maja Matari{\'c}}.}
  \bibinfo{year}{2019}\natexlab{b}.
\newblock \showarticletitle{Long-Term Personalization of an In-Home Socially
  Assistive Robot for Children With Autism Spectrum Disorders}.
\newblock \bibinfo{journal}{\emph{Frontiers in Robotics and AI}}
  \bibinfo{volume}{6} (\bibinfo{year}{2019}), \bibinfo{pages}{110}.
\newblock
\showISSN{2296-9144}
\urldef\tempurl%
\url{https://doi.org/10.3389/frobt.2019.00110}
\showDOI{\tempurl}


\bibitem[\protect\citeauthoryear{Clabaugh, Mahajan, Jain, Pakkar, Becerra, Shi,
  Deng, Lee, Ragusa, and Mataric{\'c}}{Clabaugh et~al\mbox{.}}{2019a}]%
        {cait2019frontiers}
\bibfield{author}{\bibinfo{person}{Caitlyn Clabaugh}, \bibinfo{person}{Kartik
  Mahajan}, \bibinfo{person}{Shomik Jain}, \bibinfo{person}{Roxanna Pakkar},
  \bibinfo{person}{David Becerra}, \bibinfo{person}{Zhonghao Shi},
  \bibinfo{person}{Eric Deng}, \bibinfo{person}{Rhianna Lee},
  \bibinfo{person}{Gisele Ragusa}, {and} \bibinfo{person}{Maja Mataric{\'c}}.}
  \bibinfo{year}{2019}\natexlab{a}.
\newblock \showarticletitle{Long-Term Personalization of an In-Home Socially
  Assistive Robot for Children with Autism Spectrum Disorders}.
\newblock \bibinfo{journal}{\emph{Frontiers in Robotics and AI}}
  (\bibinfo{year}{2019}).
\newblock


\bibitem[\protect\citeauthoryear{Clabaugh and Matari{\'c}}{Clabaugh and
  Matari{\'c}}{2019}]%
        {clabaugh2019escaping}
\bibfield{author}{\bibinfo{person}{Caitlyn Clabaugh} {and}
  \bibinfo{person}{Maja Matari{\'c}}.} \bibinfo{year}{2019}\natexlab{}.
\newblock \showarticletitle{Escaping oz: Autonomy in socially assistive
  robotics}.
\newblock \bibinfo{journal}{\emph{Annual Review of Control, Robotics, and
  Autonomous Systems}}  \bibinfo{volume}{2} (\bibinfo{year}{2019}),
  \bibinfo{pages}{33--61}.
\newblock


\bibitem[\protect\citeauthoryear{Coyne, Murtagh, and McGinn}{Coyne
  et~al\mbox{.}}{2020}]%
        {coyne2020dimensional}
\bibfield{author}{\bibinfo{person}{Adam~K. Coyne}, \bibinfo{person}{Andrew
  Murtagh}, {and} \bibinfo{person}{Conor McGinn}.}
  \bibinfo{year}{2020}\natexlab{}.
\newblock \showarticletitle{Using the Geneva Emotion Wheel to Measure Perceived
  Affect in Human-Robot Interaction}. In \bibinfo{booktitle}{\emph{Proceedings
  of the 2020 ACM/IEEE International Conference on Human-Robot Interaction}}
  (Cambridge, United Kingdom) \emph{(\bibinfo{series}{HRI ’20})}.
  \bibinfo{publisher}{Association for Computing Machinery},
  \bibinfo{address}{New York, NY, USA}, \bibinfo{pages}{491–498}.
\newblock
\showISBNx{9781450367462}
\urldef\tempurl%
\url{https://doi.org/10.1145/3319502.3374834}
\showDOI{\tempurl}


\bibitem[\protect\citeauthoryear{D'Mello and Graesser}{D'Mello and
  Graesser}{2011}]%
        {d2011half}
\bibfield{author}{\bibinfo{person}{Sidney D'Mello} {and} \bibinfo{person}{Art
  Graesser}.} \bibinfo{year}{2011}\natexlab{}.
\newblock \showarticletitle{The half-life of cognitive-affective states during
  complex learning}.
\newblock \bibinfo{journal}{\emph{Cognition \& Emotion}} \bibinfo{volume}{25},
  \bibinfo{number}{7} (\bibinfo{year}{2011}), \bibinfo{pages}{1299--1308}.
\newblock


\bibitem[\protect\citeauthoryear{D'Mello, Picard, and Graesser}{D'Mello
  et~al\mbox{.}}{2007}]%
        {d2007toward}
\bibfield{author}{\bibinfo{person}{Sidney D'Mello}, \bibinfo{person}{Rosalind~W
  Picard}, {and} \bibinfo{person}{Arthur Graesser}.}
  \bibinfo{year}{2007}\natexlab{}.
\newblock \showarticletitle{Toward an affect-sensitive AutoTutor}.
\newblock \bibinfo{journal}{\emph{IEEE Intelligent Systems}}
  \bibinfo{volume}{22}, \bibinfo{number}{4} (\bibinfo{year}{2007}),
  \bibinfo{pages}{53--61}.
\newblock


\bibitem[\protect\citeauthoryear{Feil-Seifer and Matari{\'c}}{Feil-Seifer and
  Matari{\'c}}{2005}]%
        {feil_defining}
\bibfield{author}{\bibinfo{person}{David Feil-Seifer} {and}
  \bibinfo{person}{Maja~J Matari{\'c}}.} \bibinfo{year}{2005}\natexlab{}.
\newblock \showarticletitle{Defining socially assistive robotics}. In
  \bibinfo{booktitle}{\emph{Rehabilitation Robotics, 2005. ICORR 2005. 9th
  International Conference on}}. IEEE, \bibinfo{pages}{465--468}.
\newblock


\bibitem[\protect\citeauthoryear{Greczek, Short, Clabaugh, Swift-Spong, and
  Mataric{\'c}}{Greczek et~al\mbox{.}}{2014}]%
        {greczek2014socially}
\bibfield{author}{\bibinfo{person}{Jillian Greczek}, \bibinfo{person}{Elaine
  Short}, \bibinfo{person}{Caitlyn~E Clabaugh}, \bibinfo{person}{Katelyn
  Swift-Spong}, {and} \bibinfo{person}{Maja Mataric{\'c}}.}
  \bibinfo{year}{2014}\natexlab{}.
\newblock \showarticletitle{Socially assistive robotics for personalized
  education for children}. In \bibinfo{booktitle}{\emph{2014 AAAI Fall
  Symposium Series}}.
\newblock


\bibitem[\protect\citeauthoryear{Gunes, Celiktutan, and Sariyanidi}{Gunes
  et~al\mbox{.}}{2019}]%
        {gunes2019live}
\bibfield{author}{\bibinfo{person}{Hatice Gunes}, \bibinfo{person}{Oya
  Celiktutan}, {and} \bibinfo{person}{Evangelos Sariyanidi}.}
  \bibinfo{year}{2019}\natexlab{}.
\newblock \showarticletitle{Live human--robot interactive public demonstrations
  with automatic emotion and personality prediction}.
\newblock \bibinfo{journal}{\emph{Philosophical Transactions of the Royal
  Society B}} \bibinfo{volume}{374}, \bibinfo{number}{1771}
  (\bibinfo{year}{2019}), \bibinfo{pages}{20180026}.
\newblock


\bibitem[\protect\citeauthoryear{Gunes and Pantic}{Gunes and Pantic}{2010}]%
        {gunes2010automatic}
\bibfield{author}{\bibinfo{person}{Hatice Gunes} {and} \bibinfo{person}{Maja
  Pantic}.} \bibinfo{year}{2010}\natexlab{}.
\newblock \showarticletitle{Automatic, dimensional and continuous emotion
  recognition}.
\newblock \bibinfo{journal}{\emph{International Journal of Synthetic Emotions
  (IJSE)}} \bibinfo{volume}{1}, \bibinfo{number}{1} (\bibinfo{year}{2010}),
  \bibinfo{pages}{68--99}.
\newblock


\bibitem[\protect\citeauthoryear{Heidgerken, Geffken, Modi, and
  Frakey}{Heidgerken et~al\mbox{.}}{2005}]%
        {heidgerken2005survey}
\bibfield{author}{\bibinfo{person}{Amanda~D Heidgerken}, \bibinfo{person}{Gary
  Geffken}, \bibinfo{person}{Avani Modi}, {and} \bibinfo{person}{Laura
  Frakey}.} \bibinfo{year}{2005}\natexlab{}.
\newblock \showarticletitle{A survey of autism knowledge in a health care
  setting}.
\newblock \bibinfo{journal}{\emph{Journal of Autism and Developmental
  disorders}} \bibinfo{volume}{35}, \bibinfo{number}{3} (\bibinfo{year}{2005}),
  \bibinfo{pages}{323--330}.
\newblock


\bibitem[\protect\citeauthoryear{Huang and Mutlu}{Huang and Mutlu}{2014}]%
        {huang2014affect-awareness}
\bibfield{author}{\bibinfo{person}{Chien-Ming Huang} {and}
  \bibinfo{person}{Bilge Mutlu}.} \bibinfo{year}{2014}\natexlab{}.
\newblock \showarticletitle{Learning-Based Modeling of Multimodal Behaviors for
  Humanlike Robots}. In \bibinfo{booktitle}{\emph{Proceedings of the 2014
  ACM/IEEE International Conference on Human-Robot Interaction}} (Bielefeld,
  Germany) \emph{(\bibinfo{series}{HRI ’14})}.
  \bibinfo{publisher}{Association for Computing Machinery},
  \bibinfo{address}{New York, NY, USA}, \bibinfo{pages}{57–64}.
\newblock
\showISBNx{9781450326582}
\urldef\tempurl%
\url{https://doi.org/10.1145/2559636.2559668}
\showDOI{\tempurl}


\bibitem[\protect\citeauthoryear{Ismail, Verhoeven, Dambre, and Wyffels}{Ismail
  et~al\mbox{.}}{2019}]%
        {ismail2019leveraging}
\bibfield{author}{\bibinfo{person}{Luthffi~Idzhar Ismail},
  \bibinfo{person}{Thibault Verhoeven}, \bibinfo{person}{Joni Dambre}, {and}
  \bibinfo{person}{Francis Wyffels}.} \bibinfo{year}{2019}\natexlab{}.
\newblock \showarticletitle{Leveraging robotics research for children with
  autism: a review}.
\newblock \bibinfo{journal}{\emph{International Journal of Social Robotics}}
  \bibinfo{volume}{11}, \bibinfo{number}{3} (\bibinfo{year}{2019}),
  \bibinfo{pages}{389--410}.
\newblock


\bibitem[\protect\citeauthoryear{Jain, Thiagarajan, Shi, Clabaugh, and
  Matari{\'c}}{Jain et~al\mbox{.}}{2020}]%
        {jain2020modeling}
\bibfield{author}{\bibinfo{person}{Shomik Jain},
  \bibinfo{person}{Balasubramanian Thiagarajan}, \bibinfo{person}{Zhonghao
  Shi}, \bibinfo{person}{Caitlyn Clabaugh}, {and} \bibinfo{person}{Maja~J
  Matari{\'c}}.} \bibinfo{year}{2020}\natexlab{}.
\newblock \showarticletitle{Modeling engagement in long-term, in-home socially
  assistive robot interventions for children with autism spectrum disorders}.
\newblock \bibinfo{journal}{\emph{Science Robotics}} \bibinfo{volume}{5},
  \bibinfo{number}{39} (\bibinfo{year}{2020}).
\newblock


\bibitem[\protect\citeauthoryear{Javed, Lee, and Park}{Javed
  et~al\mbox{.}}{2020}]%
        {javed2020towards}
\bibfield{author}{\bibinfo{person}{Hifza Javed}, \bibinfo{person}{WonHyong
  Lee}, {and} \bibinfo{person}{Chung~Hyuk Park}.}
  \bibinfo{year}{2020}\natexlab{}.
\newblock \showarticletitle{Toward an Automated Measure of Social Engagement
  for Children With Autism Spectrum Disorder—A Personalized Computational
  Modeling Approach}.
\newblock \bibinfo{journal}{\emph{Frontiers in Robotics and AI}}
  \bibinfo{volume}{7} (\bibinfo{year}{2020}), \bibinfo{pages}{43}.
\newblock
\showISSN{2296-9144}
\urldef\tempurl%
\url{https://doi.org/10.3389/frobt.2020.00043}
\showDOI{\tempurl}


\bibitem[\protect\citeauthoryear{Kasari, Sturm, and Shih}{Kasari
  et~al\mbox{.}}{2018}]%
        {kasari2018smarter}
\bibfield{author}{\bibinfo{person}{Connie Kasari}, \bibinfo{person}{Alexandra
  Sturm}, {and} \bibinfo{person}{Wendy Shih}.} \bibinfo{year}{2018}\natexlab{}.
\newblock \showarticletitle{SMARTer approach to personalizing intervention for
  children with autism spectrum disorder}.
\newblock \bibinfo{journal}{\emph{Journal of Speech, Language, and Hearing
  Research}} \bibinfo{volume}{61}, \bibinfo{number}{11} (\bibinfo{year}{2018}),
  \bibinfo{pages}{2629--2640}.
\newblock


\bibitem[\protect\citeauthoryear{Kohavi}{Kohavi}{2001}]%
        {kohavi2001crossvalidation}
\bibfield{author}{\bibinfo{person}{Ron Kohavi}.}
  \bibinfo{year}{2001}\natexlab{}.
\newblock \showarticletitle{A Study of Cross-Validation and Bootstrap for
  Accuracy Estimation and Model Selection}.
\newblock   \bibinfo{volume}{14} (\bibinfo{date}{03} \bibinfo{year}{2001}).
\newblock


\bibitem[\protect\citeauthoryear{Kort, Reilly, and Picard}{Kort
  et~al\mbox{.}}{2001}]%
        {kort2001affective}
\bibfield{author}{\bibinfo{person}{Barry Kort}, \bibinfo{person}{Rob Reilly},
  {and} \bibinfo{person}{Rosalind~W Picard}.} \bibinfo{year}{2001}\natexlab{}.
\newblock \showarticletitle{An affective model of interplay between emotions
  and learning: Reengineering educational pedagogy-building a learning
  companion}. In \bibinfo{booktitle}{\emph{Proceedings IEEE International
  Conference on Advanced Learning Technologies}}. IEEE,
  \bibinfo{pages}{43--46}.
\newblock


\bibitem[\protect\citeauthoryear{Lala, Inoue, Milhorat, and Kawahara}{Lala
  et~al\mbox{.}}{2017}]%
        {lala2017detection}
\bibfield{author}{\bibinfo{person}{Divesh Lala}, \bibinfo{person}{Koji Inoue},
  \bibinfo{person}{Pierrick Milhorat}, {and} \bibinfo{person}{Tatsuya
  Kawahara}.} \bibinfo{year}{2017}\natexlab{}.
\newblock \showarticletitle{Detection of social signals for recognizing
  engagement in human-robot interaction}.
\newblock \bibinfo{journal}{\emph{arXiv preprint arXiv:1709.10257}}
  (\bibinfo{year}{2017}).
\newblock


\bibitem[\protect\citeauthoryear{Lepper and Chabay}{Lepper and Chabay}{1988}]%
        {lepper1988socializing}
\bibfield{author}{\bibinfo{person}{Mark~R Lepper} {and} \bibinfo{person}{Ruth~W
  Chabay}.} \bibinfo{year}{1988}\natexlab{}.
\newblock \showarticletitle{Socializing the intelligent tutor: Bringing empathy
  to computer tutors}.
\newblock In \bibinfo{booktitle}{\emph{Learning issues for intelligent tutoring
  systems}}. \bibinfo{publisher}{Springer}, \bibinfo{pages}{242--257}.
\newblock


\bibitem[\protect\citeauthoryear{Leyzberg, Ramachandran, and
  Scassellati}{Leyzberg et~al\mbox{.}}{2018}]%
        {leyzberg2018effect}
\bibfield{author}{\bibinfo{person}{Daniel Leyzberg}, \bibinfo{person}{Aditi
  Ramachandran}, {and} \bibinfo{person}{Brian Scassellati}.}
  \bibinfo{year}{2018}\natexlab{}.
\newblock \showarticletitle{The effect of personalization in longer-term robot
  tutoring}.
\newblock \bibinfo{journal}{\emph{ACM Transactions on Human-Robot Interaction
  (THRI)}} \bibinfo{volume}{7}, \bibinfo{number}{3} (\bibinfo{year}{2018}),
  \bibinfo{pages}{1--19}.
\newblock


\bibitem[\protect\citeauthoryear{Matari{\'c} and Scassellati}{Matari{\'c} and
  Scassellati}{2016}]%
        {mataric_sar_handbook}
\bibfield{author}{\bibinfo{person}{Maja~J Matari{\'c}} {and}
  \bibinfo{person}{Brian Scassellati}.} \bibinfo{year}{2016}\natexlab{}.
\newblock \showarticletitle{Socially assistive robotics}.
\newblock In \bibinfo{booktitle}{\emph{Springer Handbook of Robotics}}.
  \bibinfo{publisher}{Springer}, \bibinfo{pages}{1973--1994}.
\newblock


\bibitem[\protect\citeauthoryear{Pakkar, Clabaugh, Lee, Deng, and
  Mataric{\'c}}{Pakkar et~al\mbox{.}}{2019}]%
        {pakkar2019designing}
\bibfield{author}{\bibinfo{person}{Roxanna Pakkar}, \bibinfo{person}{Caitlyn
  Clabaugh}, \bibinfo{person}{Rhianna Lee}, \bibinfo{person}{Eric Deng}, {and}
  \bibinfo{person}{Maja~J Mataric{\'c}}.} \bibinfo{year}{2019}\natexlab{}.
\newblock \showarticletitle{Designing a Socially Assistive Robot for Long-Term
  In-Home Use for Children with Autism Spectrum Disorders}. In
  \bibinfo{booktitle}{\emph{2019 28th IEEE International Conference on Robot
  and Human Interactive Communication (RO-MAN)}}. IEEE, \bibinfo{pages}{1--7}.
\newblock


\bibitem[\protect\citeauthoryear{Paulmann, Bleichner, and Kotz}{Paulmann
  et~al\mbox{.}}{2013}]%
        {paulmann2013valencearousal}
\bibfield{author}{\bibinfo{person}{Silke Paulmann}, \bibinfo{person}{Martin
  Bleichner}, {and} \bibinfo{person}{Sonja Kotz}.}
  \bibinfo{year}{2013}\natexlab{}.
\newblock \showarticletitle{Valence, arousal, and task effects in emotional
  prosody processing}.
\newblock \bibinfo{journal}{\emph{Frontiers in Psychology}}
  \bibinfo{volume}{4} (\bibinfo{year}{2013}), \bibinfo{pages}{345}.
\newblock
\showISSN{1664-1078}
\urldef\tempurl%
\url{https://doi.org/10.3389/fpsyg.2013.00345}
\showDOI{\tempurl}


\bibitem[\protect\citeauthoryear{Pedregosa, Varoquaux, Gramfort, Michel,
  Thirion, Grisel, Blondel, Prettenhofer, Weiss, Dubourg,
  et~al\mbox{.}}{Pedregosa et~al\mbox{.}}{2011}]%
        {pedregosa2011scikit}
\bibfield{author}{\bibinfo{person}{Fabian Pedregosa}, \bibinfo{person}{Ga{\"e}l
  Varoquaux}, \bibinfo{person}{Alexandre Gramfort}, \bibinfo{person}{Vincent
  Michel}, \bibinfo{person}{Bertrand Thirion}, \bibinfo{person}{Olivier
  Grisel}, \bibinfo{person}{Mathieu Blondel}, \bibinfo{person}{Peter
  Prettenhofer}, \bibinfo{person}{Ron Weiss}, \bibinfo{person}{Vincent
  Dubourg}, {et~al\mbox{.}}} \bibinfo{year}{2011}\natexlab{}.
\newblock \showarticletitle{Scikit-learn: Machine learning in Python}.
\newblock \bibinfo{journal}{\emph{the Journal of machine Learning research}}
  \bibinfo{volume}{12} (\bibinfo{year}{2011}), \bibinfo{pages}{2825--2830}.
\newblock


\bibitem[\protect\citeauthoryear{Picard}{Picard}{2000}]%
        {picard2000affective}
\bibfield{author}{\bibinfo{person}{Rosalind~W Picard}.}
  \bibinfo{year}{2000}\natexlab{}.
\newblock \bibinfo{booktitle}{\emph{Affective computing}}.
\newblock


\bibitem[\protect\citeauthoryear{Quigley, Conley, Gerkey, Faust, Foote, Leibs,
  Wheeler, and Ng}{Quigley et~al\mbox{.}}{2009}]%
        {quigley2009ros}
\bibfield{author}{\bibinfo{person}{Morgan Quigley}, \bibinfo{person}{Ken
  Conley}, \bibinfo{person}{Brian Gerkey}, \bibinfo{person}{Josh Faust},
  \bibinfo{person}{Tully Foote}, \bibinfo{person}{Jeremy Leibs},
  \bibinfo{person}{Rob Wheeler}, {and} \bibinfo{person}{Andrew~Y Ng}.}
  \bibinfo{year}{2009}\natexlab{}.
\newblock \showarticletitle{ROS: an open-source Robot Operating System}. In
  \bibinfo{booktitle}{\emph{ICRA workshop on open source software}},
  Vol.~\bibinfo{volume}{3}. Kobe, Japan, \bibinfo{pages}{5}.
\newblock


\bibitem[\protect\citeauthoryear{Rudovic, Lee, Dai, Schuller, and
  Picard}{Rudovic et~al\mbox{.}}{2018a}]%
        {rudovic2018personalized}
\bibfield{author}{\bibinfo{person}{Ognjen Rudovic}, \bibinfo{person}{Jaeryoung
  Lee}, \bibinfo{person}{Miles Dai}, \bibinfo{person}{Bj{\"o}rn Schuller},
  {and} \bibinfo{person}{Rosalind~W Picard}.} \bibinfo{year}{2018}\natexlab{a}.
\newblock \showarticletitle{Personalized machine learning for robot perception
  of affect and engagement in autism therapy}.
\newblock \bibinfo{journal}{\emph{Science Robotics}} \bibinfo{volume}{3},
  \bibinfo{number}{19} (\bibinfo{year}{2018}).
\newblock


\bibitem[\protect\citeauthoryear{Rudovic, Lee, Mascarell-Maricic, Schuller, and
  Picard}{Rudovic et~al\mbox{.}}{2017}]%
        {Rudovic2017}
\bibfield{author}{\bibinfo{person}{Ognjen Rudovic}, \bibinfo{person}{Jaeryoung
  Lee}, \bibinfo{person}{Lea Mascarell-Maricic}, \bibinfo{person}{Bj{\"o}rn~W.
  Schuller}, {and} \bibinfo{person}{Rosalind~W. Picard}.}
  \bibinfo{year}{2017}\natexlab{}.
\newblock \showarticletitle{Measuring Engagement in Robot-Assisted Autism
  Therapy: A Cross-Cultural Study}.
\newblock \bibinfo{journal}{\emph{Frontiers in Robotics and AI}}
  \bibinfo{volume}{4} (\bibinfo{year}{2017}), \bibinfo{pages}{36}.
\newblock
\showISSN{2296-9144}
\urldef\tempurl%
\url{https://doi.org/10.3389/frobt.2017.00036}
\showDOI{\tempurl}


\bibitem[\protect\citeauthoryear{Rudovic, Park, Busche, Schuller, Breazeal, and
  Picard}{Rudovic et~al\mbox{.}}{2019}]%
        {rudovic2019personalized}
\bibfield{author}{\bibinfo{person}{Ognjen Rudovic}, \bibinfo{person}{Hae~Won
  Park}, \bibinfo{person}{John Busche}, \bibinfo{person}{Bj{\"o}rn Schuller},
  \bibinfo{person}{Cynthia Breazeal}, {and} \bibinfo{person}{Rosalind~W
  Picard}.} \bibinfo{year}{2019}\natexlab{}.
\newblock \showarticletitle{Personalized estimation of engagement from videos
  using active learning with deep reinforcement learning}. In
  \bibinfo{booktitle}{\emph{2019 IEEE Conference on Computer Vision and Pattern
  Recognition Workshops (CVPR'W)}}. IEEE, \bibinfo{pages}{217--226}.
\newblock


\bibitem[\protect\citeauthoryear{Rudovic, Utsumi, Lee, Hernandez, Ferrer,
  Schuller, and Picard}{Rudovic et~al\mbox{.}}{2018b}]%
        {rudovic2018culturenet}
\bibfield{author}{\bibinfo{person}{Ognjen Rudovic}, \bibinfo{person}{Yuria
  Utsumi}, \bibinfo{person}{Jaeryoung Lee}, \bibinfo{person}{Javier Hernandez},
  \bibinfo{person}{Eduardo~Castell{\'o} Ferrer}, \bibinfo{person}{Bj{\"o}rn
  Schuller}, {and} \bibinfo{person}{Rosalind~W Picard}.}
  \bibinfo{year}{2018}\natexlab{b}.
\newblock \showarticletitle{Culturenet: A deep learning approach for engagement
  intensity estimation from face images of children with autism}. In
  \bibinfo{booktitle}{\emph{2018 IEEE/RSJ International Conference on
  Intelligent Robots and Systems (IROS)}}. IEEE, \bibinfo{pages}{339--346}.
\newblock


\bibitem[\protect\citeauthoryear{Sanghvi, Castellano, Leite, Pereira, McOwan,
  and Paiva}{Sanghvi et~al\mbox{.}}{2011}]%
        {sanghvi2011automatic}
\bibfield{author}{\bibinfo{person}{Jyotirmay Sanghvi}, \bibinfo{person}{Ginevra
  Castellano}, \bibinfo{person}{Iolanda Leite}, \bibinfo{person}{Andr{\'e}
  Pereira}, \bibinfo{person}{Peter~W McOwan}, {and} \bibinfo{person}{Ana
  Paiva}.} \bibinfo{year}{2011}\natexlab{}.
\newblock \showarticletitle{Automatic analysis of affective postures and body
  motion to detect engagement with a game companion}. In
  \bibinfo{booktitle}{\emph{Proceedings of the 6th international conference on
  Human-robot interaction}}. \bibinfo{pages}{305--312}.
\newblock


\bibitem[\protect\citeauthoryear{Scassellati, Admoni, and
  Matari{\'c}}{Scassellati et~al\mbox{.}}{2012}]%
        {scassellati2012robots}
\bibfield{author}{\bibinfo{person}{Brian Scassellati}, \bibinfo{person}{Henny
  Admoni}, {and} \bibinfo{person}{Maja Matari{\'c}}.}
  \bibinfo{year}{2012}\natexlab{}.
\newblock \showarticletitle{Robots for use in autism research}.
\newblock \bibinfo{journal}{\emph{Annual review of biomedical engineering}}
  \bibinfo{volume}{14} (\bibinfo{year}{2012}).
\newblock


\bibitem[\protect\citeauthoryear{Scassellati, Boccanfuso, Huang, Mademtzi, Qin,
  Salomons, Ventola, and Shic}{Scassellati et~al\mbox{.}}{2018}]%
        {scassellati2018improving}
\bibfield{author}{\bibinfo{person}{Brian Scassellati}, \bibinfo{person}{Laura
  Boccanfuso}, \bibinfo{person}{Chien-Ming Huang}, \bibinfo{person}{Marilena
  Mademtzi}, \bibinfo{person}{Meiying Qin}, \bibinfo{person}{Nicole Salomons},
  \bibinfo{person}{Pamela Ventola}, {and} \bibinfo{person}{Frederick Shic}.}
  \bibinfo{year}{2018}\natexlab{}.
\newblock \showarticletitle{Improving social skills in children with ASD using
  a long-term, in-home social robot}.
\newblock \bibinfo{journal}{\emph{Science Robotics}} \bibinfo{volume}{3},
  \bibinfo{number}{21} (\bibinfo{year}{2018}).
\newblock


\bibitem[\protect\citeauthoryear{Short, Swift-Spong, Greczek, Ramachandran,
  Litoiu, Grigore, Feil-Seifer, Shuster, Lee, Huang, Levonisova, Litz, Li,
  Ragusa, Spruijt-Metz, Matariánd, and Scassellati}{Short
  et~al\mbox{.}}{2014}]%
        {short2014variaty}
\bibfield{author}{\bibinfo{person}{Elaine Short}, \bibinfo{person}{Katelyn
  Swift-Spong}, \bibinfo{person}{Jillian Greczek}, \bibinfo{person}{Aditi
  Ramachandran}, \bibinfo{person}{Alexandru Litoiu}, \bibinfo{person}{Elena
  Grigore}, \bibinfo{person}{David Feil-Seifer}, \bibinfo{person}{Samuel
  Shuster}, \bibinfo{person}{Jin Lee}, \bibinfo{person}{Shaobo Huang},
  \bibinfo{person}{Svetlana Levonisova}, \bibinfo{person}{Sarah Litz},
  \bibinfo{person}{Jamy Li}, \bibinfo{person}{Gisele Ragusa},
  \bibinfo{person}{Donna Spruijt-Metz}, \bibinfo{person}{Maja Matariánd},
  {and} \bibinfo{person}{Brian Scassellati}.} \bibinfo{year}{2014}\natexlab{}.
\newblock \showarticletitle{How to Train Your DragonBot: Socially Assistive
  Robots for Teaching Children About Nutrition Through Play}.
\newblock \bibinfo{journal}{\emph{Proceedings - IEEE International Workshop on
  Robot and Human Interactive Communication}}  \bibinfo{volume}{2014}.
\newblock
\urldef\tempurl%
\url{https://doi.org/10.1109/ROMAN.2014.6926371}
\showDOI{\tempurl}


\bibitem[\protect\citeauthoryear{Spaulding and Breazeal}{Spaulding and
  Breazeal}{2019}]%
        {spaulding2019frustratingly}
\bibfield{author}{\bibinfo{person}{Samuel Spaulding} {and}
  \bibinfo{person}{Cynthia Breazeal}.} \bibinfo{year}{2019}\natexlab{}.
\newblock \showarticletitle{Frustratingly Easy Personalization for Real-Time
  Affect Interpretation of Facial Expression}. In
  \bibinfo{booktitle}{\emph{2019 8th International Conference on Affective
  Computing and Intelligent Interaction (ACII)}}. IEEE,
  \bibinfo{pages}{531--537}.
\newblock


\bibitem[\protect\citeauthoryear{Spaulding, Gordon, and Breazeal}{Spaulding
  et~al\mbox{.}}{2016}]%
        {spaulding2016affect}
\bibfield{author}{\bibinfo{person}{Samuel Spaulding}, \bibinfo{person}{Goren
  Gordon}, {and} \bibinfo{person}{Cynthia Breazeal}.}
  \bibinfo{year}{2016}\natexlab{}.
\newblock \showarticletitle{Affect-aware student models for robot tutors}. In
  \bibinfo{booktitle}{\emph{Proceedings of the 2016 International Conference on
  Autonomous Agents \& Multiagent Systems}}. \bibinfo{pages}{864--872}.
\newblock


\bibitem[\protect\citeauthoryear{Stewart, Russo, Banks, Miller, and
  Burack}{Stewart et~al\mbox{.}}{2009}]%
        {stewart2009sensory}
\bibfield{author}{\bibinfo{person}{Mary~E Stewart}, \bibinfo{person}{Natalie
  Russo}, \bibinfo{person}{Jennifer Banks}, \bibinfo{person}{Louisa Miller},
  {and} \bibinfo{person}{Jacob~A Burack}.} \bibinfo{year}{2009}\natexlab{}.
\newblock \showarticletitle{Sensory characteristics in ASD}.
\newblock \bibinfo{journal}{\emph{McGill Journal of Medicine: MJM}}
  \bibinfo{volume}{12}, \bibinfo{number}{2} (\bibinfo{year}{2009}).
\newblock


\bibitem[\protect\citeauthoryear{Sun, Feng, and Saenko}{Sun
  et~al\mbox{.}}{2015}]%
        {sun2015return}
\bibfield{author}{\bibinfo{person}{Baochen Sun}, \bibinfo{person}{Jiashi Feng},
  {and} \bibinfo{person}{Kate Saenko}.} \bibinfo{year}{2015}\natexlab{}.
\newblock \showarticletitle{Return of frustratingly easy domain adaptation}.
\newblock \bibinfo{journal}{\emph{arXiv preprint arXiv:1511.05547}}
  (\bibinfo{year}{2015}).
\newblock


\bibitem[\protect\citeauthoryear{Wilson and Cook}{Wilson and Cook}{2020}]%
        {wilson2020survey}
\bibfield{author}{\bibinfo{person}{Garrett Wilson} {and}
  \bibinfo{person}{Diane~J. Cook}.} \bibinfo{year}{2020}\natexlab{}.
\newblock \showarticletitle{A Survey of Unsupervised Deep Domain Adaptation}.
\newblock \bibinfo{journal}{\emph{ACM Trans. Intell. Syst. Technol.}}
  \bibinfo{volume}{11}, \bibinfo{number}{5}, Article \bibinfo{articleno}{51}
  (\bibinfo{date}{July} \bibinfo{year}{2020}), \bibinfo{numpages}{46}~pages.
\newblock
\showISSN{2157-6904}
\urldef\tempurl%
\url{https://doi.org/10.1145/3400066}
\showDOI{\tempurl}


\bibitem[\protect\citeauthoryear{Woolf, Burleson, Arroyo, Dragon, Cooper, and
  Picard}{Woolf et~al\mbox{.}}{2009}]%
        {woolf2009affect}
\bibfield{author}{\bibinfo{person}{Beverly Woolf}, \bibinfo{person}{Winslow
  Burleson}, \bibinfo{person}{Ivon Arroyo}, \bibinfo{person}{Toby Dragon},
  \bibinfo{person}{David Cooper}, {and} \bibinfo{person}{Rosalind Picard}.}
  \bibinfo{year}{2009}\natexlab{}.
\newblock \showarticletitle{Affect-aware tutors: Recognizing and responding to
  student affect}.
\newblock \bibinfo{journal}{\emph{IJLT}}  \bibinfo{volume}{4}
  (\bibinfo{date}{01} \bibinfo{year}{2009}), \bibinfo{pages}{129--164}.
\newblock
\urldef\tempurl%
\url{https://doi.org/10.1504/IJLT.2009.028804}
\showDOI{\tempurl}


\bibitem[\protect\citeauthoryear{Zen, Sangineto, Ricci, and Sebe}{Zen
  et~al\mbox{.}}{2014}]%
        {zen2014unsupervised}
\bibfield{author}{\bibinfo{person}{Gloria Zen}, \bibinfo{person}{Enver
  Sangineto}, \bibinfo{person}{Elisa Ricci}, {and} \bibinfo{person}{Nicu
  Sebe}.} \bibinfo{year}{2014}\natexlab{}.
\newblock \showarticletitle{Unsupervised domain adaptation for personalized
  facial emotion recognition}. In \bibinfo{booktitle}{\emph{Proceedings of the
  16th international conference on multimodal interaction}}.
  \bibinfo{pages}{128--135}.
\newblock


\end{thebibliography}

\appendix
% \newpage
% \section*{Appendix}
% \subsection{Table of Comparison between Supervised and Unsupervised Domain Adaptation}

\newpage
\pagenumbering{arabic}
\setcounter{page}{1}
\setcounter{figure}{0}
\setcounter{table}{0}
\makeatletter 
\renewcommand{\thefigure}{S\@arabic\c@figure}
\renewcommand{\thetable}{S\@arabic\c@table}
\makeatother

\noindent {\bf A    Appendix}\\

{\bf A.1 Wilcoxon signed-rank tests for F1 scores of negative and positive classes}\\
For the predictions of negative classes, wilcoxon signed-rank tests also indicated significant increases in F1 score for negative arousal/valence between personalized and generic models (arousal: $Z$ = 1.761, $p$ = .039, $r$ = .321; valence: $Z$ = 2.499, $p$ = .006, $r$ = .456), and in F1 score for negative arousal/valence between personalized and individualized models (arousal: $Z$ = 1.874, $p$ = .03, $r$ = .342; valence: $Z$ = 2.215, $p$ = .013, $r$ = .404). Similar to the prior AUROC results, there was also no significant increases in F1 scores for negative arousal/valence indicated between generic and individualized models (arousal: $Z$ = 0.454, $p$ = .325, $r$ = .083; valence: $Z$ = 1.193, $p$ = .116, $r$ = .218).

For the predictions of positive classes, except that personalized models significantly outperformed individual models for arousal, there were no significant increases between different modeling methods. Wilcoxon signed-rank tests indicated significant increases in F1 scores for positive arousal between personalized and individualized models ($Z$ = 1.647, $p$ = .05, $r$ = .301). However, no significant increases were indicated in F1 scores for positive valence between personalized and individualized models ($Z$ = 0.511, $p$ = .305, $r$ = .093), in F1 scores for positive arousal/valence between personalized and generic models (arousal: $Z$ = 0.682, $p$ = .248, $r$ = .124; valence: $Z$ = 1.533, $p$ = .063, $r$ = .280), and in F1 scores for positive arousal/valence between individualized and generic models (arousal: $Z$ = 0.852, $p$ = .197, $r$ = .156; valence: $Z$ = 0.398, $p$ = .345, $r$ = .073).

{\bf A.2 Table of Comparison between Supervised and Unsupervised Domain Adaptation}\\
\begin{table}[h]
\centering
\begin{tabular}{c|c|cc}
 \thickhline
Task & Test Participant-ID & PER (s-DA) & PER (u-DA)\\
 \thickhline
\multirow{5}{*}{ Arousal } & P-5 & \textbf{0.91} & 0.81 \\
&P-7 & \textbf{0.92} & 0.76 \\
&P-9 &  \textbf{0.92} & 0.77 \\
&P-17 &  \textbf{0.92} & 0.73 \\
\cline{2-4}
&wAVE & \textbf{0.92} & 0.77 \\
 \thickhline
\multirow{5}{*}{ Valence }&P-5 & \textbf{0.84} & 0.72 \\
&P-7 & \textbf{0.88} & 0.77 \\
&P-9 & \textbf{0.87} & 0.66 \\
&P-17 & \textbf{0.87} & 0.73 \\
\cline{2-4}
&wAVE & \textbf{0.86} & 0.71 \\
 \thickhline
\end{tabular}

\caption{\textbf{Comparison between Supervised and Unsupervised Domain Adaptation.} Personalized models with u-DA performed poorly comparing to the personalized models with s-DA (Arousal: drop of 15\%; Valence: drop of 15\%)}
\label{tab:unsupervised}
\end{table}


% \subsection{Part 1}

\end{document}